\pdfoutput=1
\documentclass{article}
\usepackage[preprint]{neurips_2024}

\usepackage[utf8]{inputenc}
\usepackage[T1]{fontenc}
\usepackage{graphicx}
\usepackage{multirow}
\usepackage{amsmath,amssymb,amsfonts}
\usepackage{booktabs}
\usepackage{adjustbox}
\usepackage{algorithm}
\usepackage{algpseudocode}
\usepackage{hyperref}
\usepackage{url}
\usepackage{xr-hyper}
\newcommand{\Head}[1]{\section{#1}}
\newcommand{\MethodHead}[1]{\section{#1}}
\newcommand{\SubHead}[1]{\subsection{#1}}

\title{CoFINN: Conservation Flux Informed Neural Networks for Physics Problems Governed by Conservation Laws}

\author{%
  Adnan Harun Do\u{g}an\thanks{Corresponding author: \texttt{doganh@metu.edu.tr}} \\
  Dept.\ of Computer Engineering \\
  Middle East Technical University \\
  Ankara, Turkey \\
  \And
  Mert Deniz \\
  Dept.\ of Mechanical Engineering \\
  Middle East Technical University \\
  Ankara, Turkey \\
  \AND
  Hande Alemdar \\
  Dept.\ of Computer Engineering \\
  Middle East Technical University \\
  Ankara, Turkey \\
  \And
  \"Ozg\"ur U\u{g}ra\c{s} Baran \\
  Dept.\ of Mechanical Engineering \\
  Middle East Technical University \\
  Ankara, Turkey \\
}

\begin{document}
\maketitle

\begin{abstract}
We present CoFINN (Conservation Flux Informed Neural Networks), a physics-informed deep learning framework for predicting compressible flow fields governed by conservation laws. Unlike conventional data-driven convolutional neural networks (CNNs), which optimize only pixel-wise similarity metrics, CoFINN embeds finite-volume conservation physics directly into the training process. Unlike classical physics-informed methods which enforce differential-equation residuals at collocation points through automatic differentiation, CoFINN adopts a finite-volume perspective consistent with modern CFD methodology. CoFINN interprets CNN output fields as structured computational grids, where each pixel represents a finite-volume cell, and enforces conservation consistency through sophisticated numerical flux calculations. The framework is evaluated on transonic flow prediction around airfoils at ($M=0.7$, $Re=6\times10^6$), including challenging conditions involving shock waves and high angles of attack. Results show that CoFINN improves aerodynamic force prediction accuracy, reducing drag prediction error by up to 34\% at extreme angles of attack and by approximately 15\% on average across the test set. Improvements are particularly significant in limited-data regimes, demonstrating that the conservation-based loss acts as an effective physical regularizer. The proposed approach maintains the computational efficiency advantages of CNN surrogates while significantly improving physical consistency and conservation behavior. The framework is architecture-agnostic and extensible to broader classes of conservation-law-governed physical systems.
\end{abstract}

\Head{Introduction}

Traditional machine learning approaches for fluid-flow prediction have demonstrated impressive speed and interpolation capability, yet they often remain fundamentally data-driven approximators that lack awareness of the governing physics. In aerodynamic applications, this limitation becomes especially critical because physically meaningful flow behavior is governed not merely by visual similarity of flow fields, but by strict conservation laws that couple pressure, density, momentum, velocity, and energy throughout the domain.

Prediction of flow around aerodynamic bodies is central to many engineering applications, including aircraft design, turbomachinery, automotive aerodynamics, and renewable energy systems. High-fidelity Computational Fluid Dynamics (CFD) simulations can accurately capture complex phenomena such as shock waves, flow separation, and boundary-layer development, but these simulations remain computationally expensive for iterative design and optimization studies. As a result, machine learning methods have emerged as promising surrogate models capable of providing rapid approximations of detailed flow fields.
Among these approaches, Convolutional Neural Networks (CNNs) are particularly attractive because flow solutions on structured grids resemble image-like data representations. CNN-based surrogate models can therefore learn mappings between geometric or flow-condition inputs and corresponding flow-field outputs. Prior studies have demonstrated substantial computational speedups while retaining reasonable predictive accuracy for aerodynamic quantities and flow structures.
However, purely data-driven CNN models suffer from an important limitation: they are not inherently constrained by the physical laws governing fluid motion. Standard training procedures optimize pixel-wise similarity metrics without explicitly enforcing conservation of mass, momentum, or energy. Consequently, predicted flow fields may appear visually plausible while still violating the underlying physics of compressible flow.

This limitation becomes especially pronounced in regions dominated by strong physical interactions, such as shock waves, boundary layers, and separated flow regions. In such areas, small spatial inconsistencies in predicted pressure or velocity fields can lead to significant violations of conservation principles and large errors in derived engineering quantities such as lift and drag. Furthermore, conventional CNNs do not inherently preserve the physical coupling between variables such as pressure, density, velocity, and energy, even though these relationships are fundamentally governed by the Navier–Stokes equations and the equation of state.

The CNN model primarily learns to map input images to output images without explicitly understanding the underlying physics of the flow. 
This raises concerns about the physical accuracy of the predictions, even if they appear visually similar to the CFD results. 
The model lacks an inherent connection between the predicted pressure and velocity distributions, even though these quantities are fundamentally linked by physical laws. 
A significant portion of the prediction errors are concentrated around shock waves. 
The sharp gradients in pressure and velocity in these regions means that even small spatial inaccuracies in the CNN's predictions leads to large errors in standard metrics. 
For example, the prediction of shock strength can be accurate enough, but a few pixel shifts of the shock location can lead to discrepancies in force calculations since the flow properties at the both sides of the shock are very different.
Similar issues are observed near the object's surface (wall boundaries). 
The CNN model can not capture the geometric details as finely as the CFD mesh, and it is not possible to directly impose physical boundary conditions. This is especially true considering the no-slip boundary conditions at the solid walls and thin boundary-layer region.
%\end{enumerate}

These observations motivate the central question addressed in this study: ``Can a neural-network-based flow predictor be trained in a manner that directly incorporates the conservation physics governing compressible flow, rather than relying solely on data similarity?''
To address this question, we propose CoFINN: Conservation Flux Informed Neural Networks, a physics-informed learning framework in which the neural network is trained not only against CFD-generated reference fields, but also against discretized conservation-law residuals computed directly from the predicted flow field itself. 
Unlike conventional Physics-Informed Neural Networks (PINNs), which typically enforce differential-equation residuals at collocation points, the present approach embeds finite-volume conservation directly into the learning process through numerical flux calculations.
The key idea underlying this work is to reinterpret the CNN output field as a finite-volume computational mesh. 
Each pixel of the predicted flow image is treated as a control volume, while neighboring pixels define computational interfaces across which numerical fluxes are evaluated. 
This interpretation establishes a direct correspondence between modern CFD methodology and neural-network training. 
By computing intercell fluxes using a Godunov-type Harten-Lax-van Leer Contact (HLLC) Riemann solver, the model is trained to minimize violations of the same conservation principles enforced in high-fidelity CFD solvers.
This formulation transforms the neural network from a purely image-processing system into a conservation-aware predictive model. 
Rather than learning only geometric correlations in the data, the model is explicitly guided toward physically admissible solutions that satisfy discrete conservation behavior throughout the domain.

%These observations highlight a general shortcoming of purely data-driven CNN models: they process data effectively but lack an intrinsic understanding of the physical processes governing the phenomena.

%Therefore, this research addresses the question: ``Can we develop a machine learning model for flow prediction that is more inherently aware of the underlying physics of the problem by using a physics-based loss function?''

Our focus is on the prediction of compressible flow around airfoils at moderate to extreme angles of attack, a problem characterized by complex fluid physics. The fundamental principles governing fluid flow are the conservation laws: conservation of mass, momentum, and energy. For compressible flows, an equation of state, such as the ideal gas law used in our case, is also essential.
These conservation laws state that the total amount of a conserved property within a defined volume remains constant over time, unless there is a net flow (flux) of that property across the boundaries of the volume. 
Finite-volume method (FVM)-based CFD solvers directly implement these conservation laws by considering discrete computational cells (finite volumes). 
In FVM, the conservation laws are interpreted such that the sum of the fluxes through the surfaces of a cell equals the rate of change of the conserved variables within that cell.

\begin{figure}
\centering
\includegraphics[width=\textwidth]{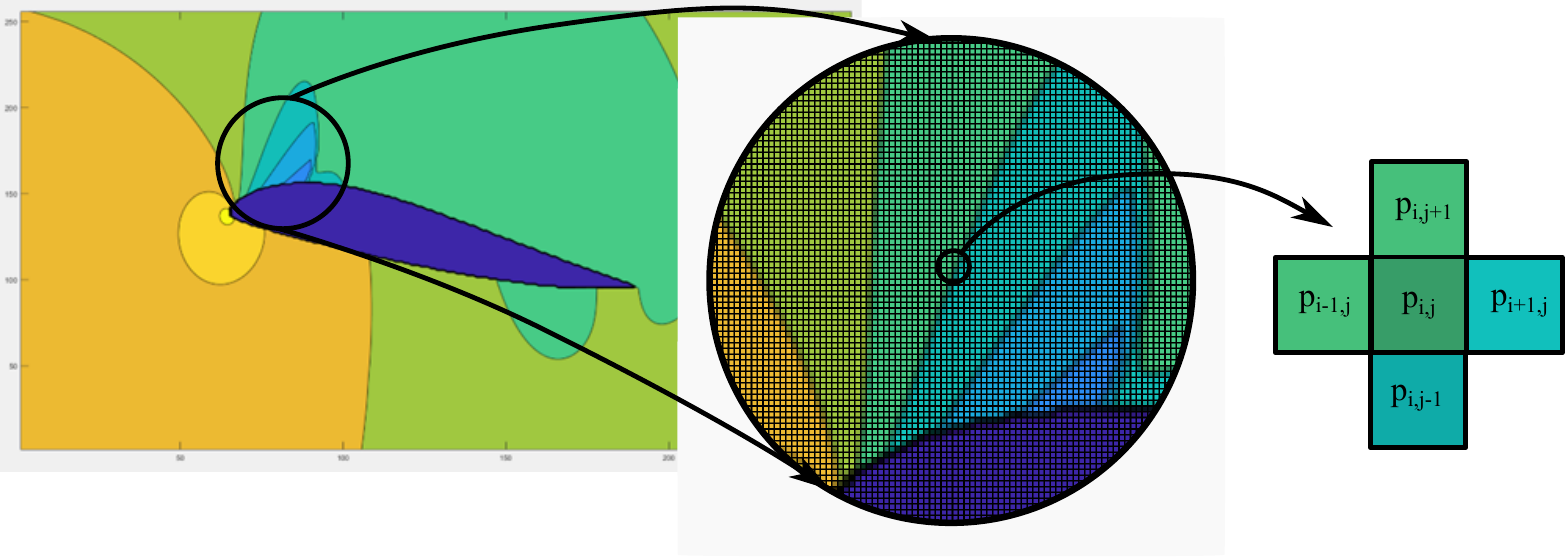}
\caption{\label{fig:Pixels-as-CV}Pixels as control volumes concept}
\end{figure}

We propose interpreting a CNN's output ``image'' as a computational grid, where each pixel represents a computational cell, and the interfaces between pixels can be seen as the control surfaces through which fluxes occur. Figure \ref{fig:Pixels-as-CV} shows a typical pressure field around an airfoil on the left. Pixels of the image essentially form a structured grid. Each cell, then, can be assigned to the pressure at the center of the cell. The grid connections are automatically provided within this embedded grid as depicted in the rightmost sketch in Figure \ref{fig:Pixels-as-CV}. Similarly ``cell values'' of other primitive (i.e. density velocity, temperature) or conservative variables (i.e. mass, momentum, energy) can be assigned from the available ``image'' data. Therefore, conservation laws can be applied on each Finite Volume Cell shown right. 

This allows us to train our model not only to match the provided CFD data but also to minimize the errors in the conserved physical quantities using calculated fluxes across these ``cell interfaces.'' Our hypothesis is that the CoFINN loss will lead to predictions with comparable accuracy to purely data-driven models but with significantly improved physical fidelity. This, in turn, should result in better usability and reliability of the predictions for real-world design applications.
To achieve this, we have implemented a detailed and carefully designed approach. Our high-fidelity CFD data was generated using a flux calculation scheme for compressible flows. We have selected the HLLC flux scheme \cite{Toro1994TheDO}, which is considered highly accurate and robust method. The HLLC scheme is a sophisticated approximate Riemann solver suitable for a wide range of flow speeds, from low subsonic to very high supersonic regimes. We have also utilized the same advanced HLLC flux scheme to calculate the CoFINN loss within our CNN model. Note that the implementation of the model is not depend on any flux calculation method, nor it requires the same method for the flux calculation method of CFD and loss function calculations. the method devised in this study can be generalized to any flux calculation method pairs.

By incorporating the CoFINN loss, we aim to produce flow field predictions around airfoils that exhibit better physical consistency. We anticipate that the predicted pressure and velocity fields will be more intrinsically linked. While we might observe a slight decrease in the overall accuracy as measured by traditional pixel-wise error metrics for individual flow variables, we expect significant improvements in the prediction of physically derived quantities such as forces and a more accurate representation of critical flow features like shock waves. 

Machine learning for aerodynamic prediction has progressed from early neural network approaches for coefficient estimation~\cite{Faller1996NeuralNA, Rai2000AerodynamicDesignNN, thirumalainambi2003training, Suresh2003LiftCP} to modern architectures capable of predicting full spatial flow fields. CNNs have been particularly successful for flow prediction tasks~\cite{Guo2016FlowApprox, Singh2017MLturbulent, Yilmaz2017ACN, Jin2018Prediction, Fukami2019SuperResolution, lee2019data, Viquerat2020SupervisedNN}, demonstrating effectiveness for steady and unsteady flows, turbulent separated flows, super-resolution reconstruction, and drag prediction. 
Encoder-decoder and U-Net architectures~\cite{chen2019unet, Thuerey2020, Bhatnagar2019, Sekar2019, Duru2020, duru_deep_2022, ruiling_deep_2025} leverage skip connections to preserve spatial information across resolution levels, enabling predictions of pressure and velocity fields around airfoils with speedups of up to three orders over CFD. Duru et al.~\cite{Duru2020, duru_deep_2022} extended these approaches to compressible and transonic regimes with the CNNFOIL architecture.

%Prior work on machine-learning surrogates for compressible airfoil flow has built primarily on U-Net architectures~\cite{chen2019unet, Thuerey2020, Bhatnagar2019, Duru2020, duru_deep_2022}, with extensions to transonic regimes~\cite{chen_towards_2023, chen_predictive_transonic_2024} and to Fourier neural operators~\cite{wen_u-fno_2022, qin_toward_2024} or graph networks~\cite{Catalani2023-ps, nabian_x-meshgraphnet_2024} for unstructured grids. Physics-informed approaches~\cite{sharma_review_2023, bastek_physics-informed_2024, hu_generative_2024} penalize PDE residuals at collocation points but do not enforce conservation in a finite-volume sense. Recent work~\cite{patsatzis_gorinns_2025} combines Godunov-type fluxes with shallow networks for 1D inverse problems. The present work is the first to embed full HLLC flux supervision directly into 2D CNN-class architectures and validate it across four families (CNN, FNO, ViT, diffusion).

Specialized architectures have been developed for challenging compressible and transonic flow regimes~\cite{hui_fast_2020, Wang2021, WU2022, chen_towards_2023, chen_predictive_transonic_2024, wang_general_2023, Zuo2022-ow, Catalani2023-ps}, addressing shock wave prediction, high Reynolds number flows, and data scarcity. Chen and Thuerey~\cite{chen_towards_2023, chen_predictive_transonic_2024} achieved order-of-magnitude improvements through canonical space encoding and attention-based U-Net models with roll-out training for unsteady transonic flows, while Catalani et al.~\cite{Catalani2023-ps} showed that Graph Neural Networks outperform CNNs for shock capture on unstructured meshes. Graph Neural Networks have emerged as powerful tools for handling unstructured meshes~\cite{nabian_x-meshgraphnet_2024, Catalani2023-ps}, while neural operators, particularly the Fourier Neural Operator (FNO)~\cite{wen_u-fno_2022, qin_toward_2024, zuo_fast_2023}, learn solution operators in Fourier space with resolution independence. Implicit neural representations~\cite{catalani_aero-nef_2024, xiao_learning_2024} provide continuous flow field representations with five orders of magnitude speedup over traditional solvers.

Physics-Informed Neural Networks (PINNs)~\cite{sharma_review_2023} embed physical laws into the learning process by penalizing partial differential equation (PDE) residuals, with applications to interface problems and moving boundaries~\cite{zhu_physics-informed_2023, roy_adaptive_2024, mullins_physics-informed_2025, bi_extended_2025, zhu_physics-informed_2024, lu_physics-informed_2024}, airfoil pressure prediction~\cite{huang_physics-informed_2025}, transonic airfoil flows with artificial dissipation~\cite{wassing_pinns_transonic_2025}, flight dynamics~\cite{michek_flight_2024}, and specialized formulations including physics-informed extreme learning machines~\cite{ren_physics-informed_2025} and improved PINNs with small sample learning~\cite{li_improved_2023}. Physics-informed diffusion models~\cite{bastek_physics-informed_2024, hu_generative_2024} have reduced residual errors by up to two orders of magnitude, with geometry-conditioned variants demonstrating superior performance over CNNs in handling complex geometries. Hybrid approaches combining physics-informed techniques with traditional architectures~\cite{wang_deep_2021, sharpe_neuralfoil_2025, patel_accelerated_2025, teimourian_airfoil_2024, sun_deep_2021, wu_fast_2022, hassanian_deep_2022} have achieved consistent improvements; notably NeuralFoil~\cite{sharpe_neuralfoil_2025} embeds physical symmetries structurally, achieving between $8\times$ to $1{,}000\times$ speedups over XFoil \cite{XFOIL} with guaranteed extrapolation behavior.

Despite this progress, a critical gap persists: existing methods do not explicitly enforce conservation laws at the discrete computational level. PINNs penalize PDE residuals at collocation points but do not ensure that predicted fields satisfy conservation of mass, momentum, and energy in a finite volume sense---the fundamental principle underlying CFD solvers. Recent work on Godunov-Riemann Informed Neural Networks (GoRINNs)~\cite{patsatzis_gorinns_2025} combines shallow neural networks with Godunov-type finite volume schemes to learn physical flux functions for hyperbolic conservation laws, but focuses on inverse problems for 1D scalar laws rather than supervised flow field prediction. Our proposed Conservation Flux Informed Neural Network (CoFINN) addresses this gap by treating CNN output pixels as finite volume cells and calculating fluxes across cell interfaces using the same Riemann solvers employed in high-fidelity CFD. This bridges data-driven learning with rigorous numerical methods, combining the computational efficiency of CNNs with the physical consistency of finite volume methods.

%The primary contributions of this paper are as follows: 

%\begin{itemize}
%\item We introduce a new methodology to embed physical processes directly into the training of Convolutional Neural Networks for predicting detailed fields governed by physical laws. This results in what we term a Conservation Flux Informed Neural Network (CoFINN).

%\item While demonstrated on the prediction of compressible flow around airfoils, the pixel-as-control-volume framework is architecturally compatible with any structured-grid conservation law problem where an appropriate Riemann solver is available. The airfoil problem serves as a representative and challenging test case for the methodology.

%\item We explicitly differentiate our method from Physics-Informed Neural Networks (PINNs). While acknowledging the ingenuity of PINNs, we highlight the computational advantages and robustness of CNNs, particularly when high-resolution predictions of physical fields across a domain are required, where PINNs can become computationally expensive. 

%\item We present an improved method for calculating aerodynamic forces (specifically lift and drag) around the airfoil based on the CNN predictions. This addresses a common challenge in CNN-based flow predictions where resolving fine-scale details near solid walls is difficult. Our proposed method aims to circumvent this limitation and the absence of shear stress data in our current model to provide more accurate force estimations.
%\end{itemize}

\Head{Related Work}
\label{sec:related-work}

Machine learning for aerodynamic prediction has progressed from early neural network approaches for coefficient estimation~\cite{Faller1996NeuralNA, Rai2000AerodynamicDesignNN, thirumalainambi2003training, Suresh2003LiftCP} to modern architectures capable of predicting full spatial flow fields. Convolutional Neural Networks have been particularly successful for flow prediction tasks~\cite{Guo2016FlowApprox, Singh2017MLturbulent, Yilmaz2017ACN, Jin2018Prediction, Fukami2019SuperResolution, lee2019data, Viquerat2020SupervisedNN}, demonstrating effectiveness for steady and unsteady flows, turbulent separated flows, super-resolution reconstruction, and drag prediction. Encoder-decoder and U-Net architectures~\cite{chen2019unet, Thuerey2020, Bhatnagar2019, Sekar2019, Duru2020, duru_deep_2022, ruiling_deep_2025} leverage skip connections to preserve spatial information across resolution levels, enabling predictions of pressure and velocity fields around airfoils with speedups of up to three orders over CFD. Duru et al.~\cite{Duru2020, duru_deep_2022} extended these approaches to compressible and transonic regimes with the CNNFOIL architecture, upon which the present work builds.

Specialized architectures have been developed for challenging compressible and transonic flow regimes~\cite{hui_fast_2020, Wang2021, WU2022, chen_towards_2023, chen_predictive_transonic_2024, wang_general_2023, Zuo2022-ow, Catalani2023-ps}, addressing shock wave prediction, high Reynolds number flows, and data scarcity. Chen and Thuerey~\cite{chen_towards_2023, chen_predictive_transonic_2024} achieved order-of-magnitude improvements through canonical space encoding and attention-based U-Net models with roll-out training for unsteady transonic flows, while Catalani et al.~\cite{Catalani2023-ps} showed that Graph Neural Networks outperform CNNs for shock capture on unstructured meshes. Graph Neural Networks have emerged as powerful tools for handling unstructured meshes~\cite{nabian_x-meshgraphnet_2024, Catalani2023-ps}, while neural operators, particularly the Fourier Neural Operator~\cite{wen_u-fno_2022, qin_toward_2024, zuo_fast_2023}, learn solution operators in Fourier space with resolution independence. Implicit neural representations~\cite{catalani_aero-nef_2024, xiao_learning_2024} provide continuous flow field representations with five orders of magnitude speedup over traditional solvers.

Physics-Informed Neural Networks (PINNs)~\cite{sharma_review_2023} embed physical laws into the learning process by penalizing PDE residuals, with applications to interface problems and moving boundaries~\cite{zhu_physics-informed_2023, roy_adaptive_2024, mullins_physics-informed_2025, bi_extended_2025, zhu_physics-informed_2024, lu_physics-informed_2024}, airfoil pressure prediction~\cite{huang_physics-informed_2025, jiazhe_airfoil_nodate}, transonic airfoil flows with artificial dissipation~\cite{wassing_pinns_transonic_2025}, flight dynamics~\cite{michek_flight_2024}, and specialized formulations including physics-informed extreme learning machines~\cite{ren_physics-informed_2025} and improved PINNs with small sample learning~\cite{li_improved_2023}. Physics-informed diffusion models~\cite{bastek_physics-informed_2024, hu_generative_2024} have reduced residual errors by up to two orders of magnitude, with geometry-conditioned variants demonstrating superior performance over CNNs in handling complex geometries. Hybrid approaches combining physics-informed techniques with traditional architectures~\cite{wang_deep_2021, sharpe_neuralfoil_2025, patel_accelerated_2025, teimourian_airfoil_2024, sun_deep_2021, wu_fast_2022, hassanian_deep_2022} have achieved consistent improvements; notably NeuralFoil~\cite{sharpe_neuralfoil_2025} embeds physical symmetries structurally, achieving between $8\times$ to $1{,}000\times$ speedups over XFoil \cite{XFOIL} with guaranteed extrapolation behavior.

Despite this progress, a critical gap persists: existing methods do not explicitly enforce conservation laws at the discrete computational level. PINNs penalize PDE residuals at collocation points but do not ensure that predicted fields satisfy conservation of mass, momentum, and energy in a finite volume sense---the fundamental principle underlying CFD solvers. Recent work on Godunov-Riemann Informed Neural Networks (GoRINNs)~\cite{patsatzis_gorinns_2025} combines shallow neural networks with Godunov-type finite volume schemes to learn physical flux functions for hyperbolic conservation laws, but focuses on inverse problems for 1D scalar laws rather than supervised flow field prediction. Our proposed Conservation Flux Informed Neural Network (CoFINN) addresses this gap by treating CNN output pixels as finite volume cells and calculating fluxes across cell interfaces using the same Riemann solvers employed in high-fidelity CFD. This bridges data-driven learning with rigorous numerical methods, combining the computational efficiency of CNNs with the physical consistency of finite volume methods.

%\section{Methods}
For many physical processes, including fluid flow, heat and mass transfer, electric charge, and even traffic flow, conservation laws constitute our fundamental tool for comprehending and modeling the physical mechanisms of these processes. These models include one or more conserved variables and the mechanisms defining the relationship between the local quantity of a conserved variable and its transport.

A conservation law can be described as a differential equation within a domain, or it can be formulated as an integral equation on a finite control volume using Green's theorem. From a machine learning perspective, the distribution of conserved variables can be estimated by training on existing solutions. However, there is no guarantee that the estimated solutions are physically acceptable.

We theorize that, for a physical system governed by a conservation law, more physically realistic solutions can be generated by minimizing the deviation from these laws. For this purpose, we utilize the integral form of the conservation laws and ensure the deviation from the conservation laws within the whole domain is minimized. Our neural network architecture estimates the local values of physical parameters as pixels values, and each pixel is treated as a finite control volume.

In this section, we will discuss conservation laws in general. Then, we introduce a challenging problem in compressible fluid flow that includes several coupled conserved variables and involves both convection and diffusion. Finally, we explain the essence of our method and its basic theory, followed by the simplification and implementation details for the selected model problem.
\MethodHead{Conservation laws}

For a given finite control volume $\Omega$, any conservation law in integral form is described as in \eqref{eq:conservation-integral-form}.
\begin{equation}
\frac{\partial}{\partial t}\int_{\Omega}\mathbf{U}\,d\Omega+\oint_{\partial\Omega}\mathbf{F}\cdot\mathbf{n}\,dS=\int_{\Omega}\mathbf{S}\,d\Omega
\label{eq:conservation-integral-form}
\end{equation}
In this equation, $\mathbf{U}$ is the vector of conserved variables, which consists of mass, $x$-momentum, $y$-momentum, and Total Energy. $\mathbf{F}_{c}$is the flux vector.   $\mathbf{S}$ is the source term vector, which is typically zero for standard flow problems without external forces. The second term is the flux through the control surface, integrated through surface $S$ along the boundary ($d\Omega$) of the control volume ($\Omega$). 

It is possible, and often useful to split the flux vector into its convective and diffusive parts, $\mathbf{F} = \mathbf{F}_{c} +\mathbf{F}_{v}$. Here, $ \mathbf{F}_{c}$.  is the convective (inviscid) flux vector. $\mathbf{F}_{v}$ is the viscous (convective) flux vector. Then Equation \eqref{eq:conservation-integral-form} can be written is the following form: 
\begin{equation}
\frac{\partial}{\partial t}\int_{\Omega}\mathbf{U}\,d\Omega+\oint_{\partial\Omega}\left(\mathbf{F}_{c}-\mathbf{F}_{v}\right)\cdot\mathbf{n}\,dS=\int_{\Omega}\mathbf{S}\,d\Omega
\label{eq:conservation-integral-form-spl}
\end{equation}

\subsubsection*{Conservation laws for compressible fluid flow}
Equation \eqref{eq:conservation-integral-form-spl} is a rather generalized form and can be applied in a variety of conservation-laws driven problems. For a 2D compressible flow problem, the vector of conserved variables are as such:
\begin{equation}
\mathbf{U}=\begin{pmatrix}\rho\\
\rho u\\
\rho v\\
\rho E
\end{pmatrix}
\label{eq:conserved-variables}
\end{equation}
In this equation, $\rho$ is density $u,v$ are velocity components in $x$ and $y$ directions and $E$ is the total energy per unit mass given as:
\begin{equation}
E=e+\frac{1}{2}(u^{2}+v^{2})
\label{eq:total-energy}
\end{equation} 
where, $e$ is internal energy. Therefore the volume integral given in Equation \eqref{eq:conservation-integral-form-spl} of conserved variables contained within vector $\mathbf{U}$ of Equation \eqref{eq:conserved-variables} are mass, $x$ and $y$-momentum ($\rho u$, $\rho v$) and the energy within the control volume.

Convective (inviscid) Flux Vector ($\mathbf{F}_{c}$) represents the flux of conserved quantities due to the bulk motion of the fluid and pressure forces. For 2D flow, dot product in Eq. \eqref{eq:conservation-integral-form-spl} allows us to split the flux vector in its $x$ and $y$ components, $\mathbf{F}_{c,x}$ and $\mathbf{F}_{c,y}$
\begin{equation}
\mathbf{F}_{c,x}=\begin{pmatrix}\rho u\\
\rho u^{2}+p\\
\rho uv\\
\rho uH
\end{pmatrix},\quad\mathbf{F}_{c,y}=\begin{pmatrix}\rho v\\
\rho uv\\
\rho v^{2}+p\\
\rho vH
\end{pmatrix}
\label{eq:convective-flux}
\end{equation}
In this equation, $p$ is pressure, and $H$ is the total enthalpy per unit mass
\begin{equation}
H=E+\frac{p}{\rho}
\label{eq:total-enthalpy}
\end{equation}
Viscous Flux Vector ($\mathbf{F}_{v}$) represents the flux of conserved quantities due to viscous shear stresses and heat conduction. It is also written in terms of its $x$ and $y$ component.
\begin{equation}
\mathbf{F}_{v,x}=\begin{pmatrix}0\\
\tau_{xx}\\
\tau_{xy}\\
u\tau_{xx}+v\tau_{xy}-q_{x}
\end{pmatrix},\quad\mathbf{F}_{v,y}=\begin{pmatrix}0\\
\tau_{yx}\\
\tau_{yy}\\
u\tau_{yx}+v\tau_{yy}-q_{y}
\end{pmatrix}
\label{eq:viscous-flux}
\end{equation}
In this equation, $\tau_{ij}$ are the components of the viscous stress tensor, $q_{x},q_{y}$ are the components of the heat flux vector. We assumed Newtonian fluid and Fourier's Law of conductivity for this study.

The shear stress is given using the Bousinnesq approximation as
\begin{equation}
\tau_{ij}=\mu\bigg(\frac{\partial u_{i}}{\partial x_{j}}+\frac{\partial u_{j}}{\partial x_{i}}-\frac{2}{3}\frac{\partial u_{k}}{\partial x_{k}}\delta_{ij}\bigg),
\label{eq:shear-stress-tensor}
\end{equation}
For the RANS modelling, the viscosity term, $\mu$ is the combination of laminar and turbulent counterparts. The turbulent viscosities can be calculated through turbulence models, whereas the material viscosity is calculated by the Sutherland Law. For this study we used air as the working gas with the viscosity given as Eq. \eqref{eq:sutherland-law}.
\begin{equation}
\mu=\frac{1.458\times10^{-6}T^{3/2}}{T+110.4}
\label{eq:sutherland-law}
\end{equation}
where, $T$ is temperature defined in Kelvin.

The governing equations in Eq. \eqref{eq:conservation-integral-form} contain four equations and five unknowns $(\rho,p,T,u,v)$. Therefore, a fifth equation is required, which is the state equation. We use the ideal gas law in this study, which is sufficient for the flow regimes presented in this work.
\begin{equation}
p=\rho RT
\label{eq:ideal-gas}
\end{equation}

\MethodHead{Interpreting image data as a finite-volume mesh}
The model problem can be solved by an analytical or numerical method and the solutions on the domain can be used to train machine-learning algorithm. For the selected model problem given in equations \eqref{eq:conserved-variables} to \eqref{eq:ideal-gas}, the field solutions are obtained using a finite-volume-based CFD solver. As discussed earlier, the \emph{integral form of the conservation laws} are utilized to ensure the fidelity of the solution. Therefore, finite volumes to evaluate the flux and volume integrals defined in equation \eqref{eq:conservation-integral-form} are required.

\begin{figure}
\centering
\includegraphics[width=0.6\textwidth]{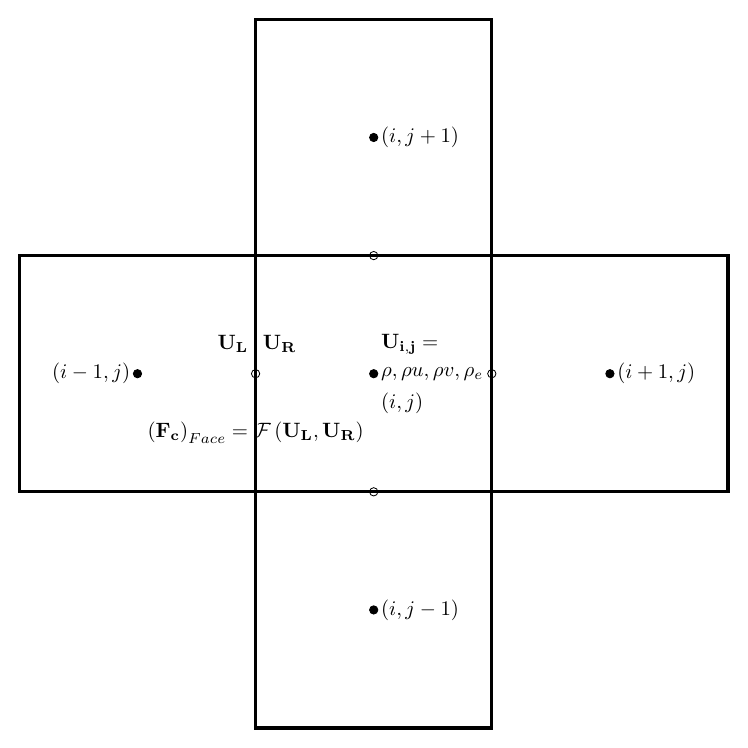}
\caption{\label{fig:CV-with-neigh}Control volume cell and its neighbors}
\end{figure}

For this purpose, the dense flow field solution obtained from the high-fidelity CFD simulation is mapped onto an ``$N\times M$ pixel array'', representing a discretized spatial domain. Within this structure, each pixel is treated as a closed, non-overlapping finite control volume (or cell), effectively transforming the image data into a computational mesh (as conceptually depicted in Figure \ref{fig:Pixels-as-CV}). Each cell is assigned a single, average value for every flow property. Specifically, the data assigned to cell $(i,j)$ is the cell-centered average of the conserved variable vector $\mathbf{U}_{i,j}$ (as shown in Figure \ref{fig:CV-with-neigh}), where $\mathbf{U}$ includes density, momentum components, and total energy.

To enforce the conservation laws, the fluxes must be evaluated across the interface (or face) connecting adjacent control volumes. For any given cell $(i,j)$, there are four surrounding interfaces where fluxes must be calculated: the north, south, east, and west faces.

At each cell interface, the constant cell-centered value must be used to define the states on either side of the face. For an interface separating cell $i$ from cell $j$, this process yields two distinct states: the left state $\mathbf{U}_{L}$ (from cell at the direction normal to the face) and the right state $\mathbf{U}_{R}$ (from the opposing cell). 

Since the governing Partial Differential Equations (PDEs) for many physical processes allow for non-smooth solutions, the left state $\mathbf{U}_{L}$ and right state $\mathbf{U}_{R}$ can be significantly different across the interface. This requires the use of a specialized numerical flux function or a Riemann solver to correctly compute the physically meaningful flux $\mathbf{F}_{k}$ across the face, ensuring the correct amount of conserved property flows between the two control volumes.

%For our specific model problem—compressible flow—we utilize the HLLC (Harten-Lax-van Leer Contact) Riemann solver. This advanced numerical scheme takes the interpolated states $\mathbf{U}_{L}$ and $\mathbf{U}_{R}$, along with the face normal vector ($\mathbf{n}$), to calculate the final convective flux $\mathbf{F}_{k}$. This approach effectively accounts for the possibility of shock waves and contact discontinuities present in transonic flow fields:
For our specific model problem --compressible flow-- we utilize the HLLC (Harten-Lax-van Leer Contact) Riemann solver. This advanced numerical scheme takes the interpolated states $\mathbf{U}_{L}$ and $\mathbf{U}_{R}$, along with the face normal vector ($\mathbf{n}$), to calculate the final convective flux $\mathbf{F}_{c}$. This approach effectively accounts for the possibility of shock waves and contact discontinuities present in transonic flow fields:
\begin{equation}
\mathbf{F}_{c}=\mathcal{F}(\mathbf{U}_{L},\mathbf{U}_{R},\mathbf{n})
\label{eq:riemann-flux}
\end{equation}

\MethodHead{Conservation Flux Informed Neural Networks, CoFINN}
\label{sec:cofinn}

To embed physical principles into our CNN training, we utilize the integral form of the conservation laws, simplifying them to focus on the steady-state and the balance between convective, viscous fluxes, and any internal source terms. Therefore, we start from the general conservation equation given in Equation \eqref{eq:conservation-integral-form-spl}.

For a steady-state flow field, the rate of change of conserved variables within a control volume $\Omega$ is zero. The general integral form of the conservation laws for steady-state simplifies to the following exact balance:
\begin{equation}
\oint_{\partial\Omega}\mathbf{F}_{c}\cdot\mathbf{n}\,dS-\oint_{\partial\Omega}\mathbf{F}_{v}\cdot\mathbf{n}\,dS=\mathbf{RHS},
\label{eq:cofinn-balance}
\end{equation}
where $\mathbf{F}_{c}$ and $\mathbf{F}_{v}$ are the convective and viscous flux vectors. Note that, splitting the fluxes into convective and diffusive parts are rather artificial and starting from equation \eqref{eq:conservation-integral-form} is also possible, if the fluxes are calculated on the pixel-based control volumes. The Right-Hand Side ($\mathbf{RHS}$) of the Equation \eqref{eq:cofinn-balance} is defined here as the net rate of external forces, work, or heat added to the control volume.

We introduce the vector $\mathbf{B}_{\text{internal}}$ to capture all source, sink, and external force effects acting within or on the control volume. This term specifically includes effects such as body forces, chemical reactions, and, crucially, solid boundary effects. For a cell adjacent to an airfoil surface, $\mathbf{B}_{\text{internal}}$ includes the net forces and work applied by the solid boundary on the fluid, such as pressure forces and work done by shear stresses at the wall.

Incorporating this term, the conservation equation becomes:

\begin{equation}
\oint_{\partial\Omega}\mathbf{F}_{c}\cdot\mathbf{n}\,dS-\oint_{\partial\Omega}\mathbf{F}_{v}\cdot\mathbf{n}\,dS=\mathbf{B}_{\text{internal}}
\label{eq:cofinn-internal}
\end{equation}
For an internal cell not connected to a solid boundary and where there are no other internal sources (like chemical reactions or dedicated heat addition), $\mathbf{B}_{\text{internal}}=\mathbf{0}$

For the purposes of our CNN training, we strategically group all non-convective terms into a single Ground Truth Flux Residual ($\mathcal{F}_{\mathrm{GT}}$), which we calculate directly from the high-fidelity CFD data. This allows us to use the prediction capability of the CNN only for the convective fluxes, while ensuring the model is accountable for the correct overall conservation balance.

We define the residual $\mathcal{F}_{\mathrm{GT}}$ for any control volume (CV) as the sum of all terms other than the convective flux contribution:
\begin{equation}
\mathcal{F}_{\mathrm{GT}}=\oint_{\partial\Omega}\mathbf{F}_{v}\cdot\mathbf{n}\,dS+\mathbf{B}_{\text{internal}}
\label{eq:gt-residual}
\end{equation}
The core conservation equation for any cell, therefore, must satisfy the following exact balance, which represents our ground truth condition:
\begin{equation}
\oint_{\partial\Omega}\mathbf{F}_{c}\cdot\mathbf{n}\,dS=\mathcal{F}_{\mathrm{GT}}
\label{eq:core-conservation}
\end{equation}

This formulation holds true for all cells, regardless of whether they are internal (where $\mathbf{B}_{\text{internal}}$ is zero, and $\mathbf{F}_{v}$ is small) or boundary-adjacent (where $\mathbf{B}_{\text{internal}}$ accounts for the wall interaction and $\mathbf{F}_{v}$ is significant). This residual $\mathcal{F}_{\mathrm{GT}}$ contains all necessary source and viscous terms required for conservation law.

It is important to note that $\mathcal{F}_{\mathrm{GT}}$ is precomputed from the CFD ground truth, making the CoFINN loss a \emph{supervised} regularizer rather than a self-supervised constraint. Unlike PINNs, which penalize PDE residuals without requiring solution data, CoFINN requires the same high-fidelity training data as a purely data-driven model. the CoFINN loss reshapes the loss landscape to favor predictions with better conservation compliance, but does not reduce data requirements.

The model is then trained to minimize the error between the right-hand side, $\mathcal{F}_{\mathrm{GT}}$, and the convective flux calculated from the network's prediction, $\mathcal{F}_{\mathrm{pred}}$:
\begin{equation}
\varepsilon\sim\left|\mathcal{F}_{\mathrm{GT}}-\mathcal{F}_{\mathrm{pred}}\right|,
\label{eq:flux-error}
\end{equation}
where $\mathcal{F}_{\mathrm{pred}}=\oint_{\partial\Omega}\mathbf{F}_{c}(\mathbf{U}_{\mathrm{pred}})\cdot\mathbf{n}\,dS$. By minimizing this error, we ensure that the predicted flow field, when plugged back into the conservation equation, satisfies the known conservation balance (which includes all boundary and viscous effects) for every cell.

\SubHead{Flux Computation}
\label{sec:hllc-algorithm}

As mentioned before, we have utilized a Riemannian flux calculation routine in this study. We have selected the HLLC flux scheme for this purpose. The implemented HLLC variant is given on the supplementary materials. Please note that, CoFINN methodology is not bound to any flux scheme. One can use any flux scheme or any discretion method. In this section we describe the flux method of our choice, that is shared between the CFD and CoFINN methods.

\subsubsection*{Pixel-Level Flux Balance}

In our image-as-mesh formulation, each pixel $(i,j)$ has four interfaces. The left and right states are obtained by picking adjacent pixel values. 
%
%(e.g., the east-face left state uses $\frac{1}{2}[\mathbf{U}_{i,j} + \mathbf{U}_{i,j+1}]$). The face normals are $\mathbf{n}_{\text{east}} = (1,0)$, $\mathbf{n}_{\text{west}} = (-1,0)$, $\mathbf{n}_{\text{north}} = (0,1)$, $\mathbf{n}_{\text{south}} = (0,-1)$, with face areas proportional to the pixel dimensions ($A_{\text{east,west}} = 1/255$, $A_{\text{north,south}} = 2/255$). 
We have used first-order discretizations, which means the left and right states at the cell face are equal to cell (pixel) value for the Riemannian flux method we use. It is also possible to use second-order accurate discretization using slope limiters. Note that if a central scheme is used, face values should be interpolated from the pixel data. 

The fluxes should be evaluated all interfaces (East, West, North, South), using the face normal. The face areas should be considered. We have used an unequal scaling in $x$ (Drag) and $y$ directions while generating the pixelated data. Therefore, we used  with face areas proportional to the pixel dimensions ($A_{\text{eas t,west}} = 1/255$, $A_{\text{north,south}} = 2/255$). The net flux imbalance for pixel $(i,j)$ is:
%
%\begin{equation}
%\varepsilon_{i,j} =  \sum_{k \in \{\text{E,W,N,S}\}} %\mathbf{F}_{\text{mom},k}  - \mathcal{F}_{\mathrm{GT},i,j}
%\label{eq:pixel-flux-balance}
%\end{equation}
\begin{equation}
\varepsilon_{i,j} = \left( \sum_{k \in \{\text{E,W,N,S}\}} \mathbf{F}_{k}\right)  - \mathcal{F}_{\mathrm{GT},i,j}
\label{eq:pixel-flux-balance}
\end{equation}
The entire CoFINN flux computation is implemented as differentiable tensor operations, enabling gradient backpropagation through the selected flux calculation method during training. Algorithm 1 given in supplementary materials %Algorithm~\ref{alg:hllc}
summarizes the complete procedure.
%

%%%%%%%%%%%%%%%%%%%% TRAINING INFO

\SubHead{Loss Functions}

Both data-driven and physics-informed losses are active from the start of training, with their relative contribution controlled by the CoFINN loss weight $\lambda$.

\subsubsection*{Numerical Loss (Data-Driven)}
The Mean Absolute Error (MAE) between predicted and ground truth primitive variables:
\begin{equation}
\mathcal{L}_{\text{data}} = \frac{1}{n}\sum_{j=1}^{n}|y_j - \hat{y}_j|
\label{eq:mae-loss}
\end{equation}
where $y_j$ represents the ground truth values, $\hat{y}_j$ represents the network predictions, and $n$ is the number of data points in the flow field.

\subsubsection*{Physical Loss (Physics-Informed)}
After computing the predicted primitive variables, we calculate momentum fluxes using the underlying flux solver of CoFINN (Eq. \eqref{eq:riemann-flux}) at each cell interface. The CoFINN loss measures the error in momentum conservation:
\begin{equation}
\mathcal{L}_{\text{flux}} = \text{MAE}(\mathcal{F}_{\text{GT}} - \mathcal{F}_{\text{pred}})
\label{eq:flux-loss}
\end{equation}
where $\mathcal{F}_{\text{pred}}$ is computed from the network's predicted flow field using the same flux calculation procedures as the CFD solver.%% (Algorithm~\ref{alg:hllc}).

The total loss function combines both components using a convex combination controlled by a single weight $\lambda \in [0, 1]$:
\begin{equation}
\mathcal{L}_{\text{total}} = (1 - \lambda) \times \mathcal{L}_{\text{data}} + \lambda \times \mathcal{L}_{\text{flux}}
\label{eq:total-loss}
\end{equation}
When $\lambda = 0$, the model reduces to a purely data-driven MAE model; when $\lambda = 1$, only the CoFINN loss is active. Intermediate values balance data fidelity with conservation compliance.

\Head{Force calculations} 
\label{sec:force-calc}

The methodology establishes a direct link between the network's output and the principles of the Finite Volume Method (FVM). As conceptually illustrated in Figure1 of main text% \ref{fig:Pixels-as-CV}
, an airfoil is positioned within the computational domain. This entire domain is represented by a large rectangular control volume (CV), which is itself composed of an $N\times M$ grid of individual cells, or ``pixels.'' Therefore, the entire image output is not merely a representation within a larger CV; rather, the entire CV is the union of the $N\times M$ discrete computational cells. Each of these cells acts as a local control volume for which the conservation laws are locally enforced. The figure depicts the conceptual flow that occurs across the external boundaries of this $N\times M$ domain, representing the interaction with the far-field conditions. This structured discretization allows us to directly map the network's pixel-based predictions to a finite volume framework, enabling the calculation of fluxes and the assessment of conservation law compliance at a granular level across the entire domain.

\begin{figure}
\centering
\includegraphics[width=0.8\textwidth]{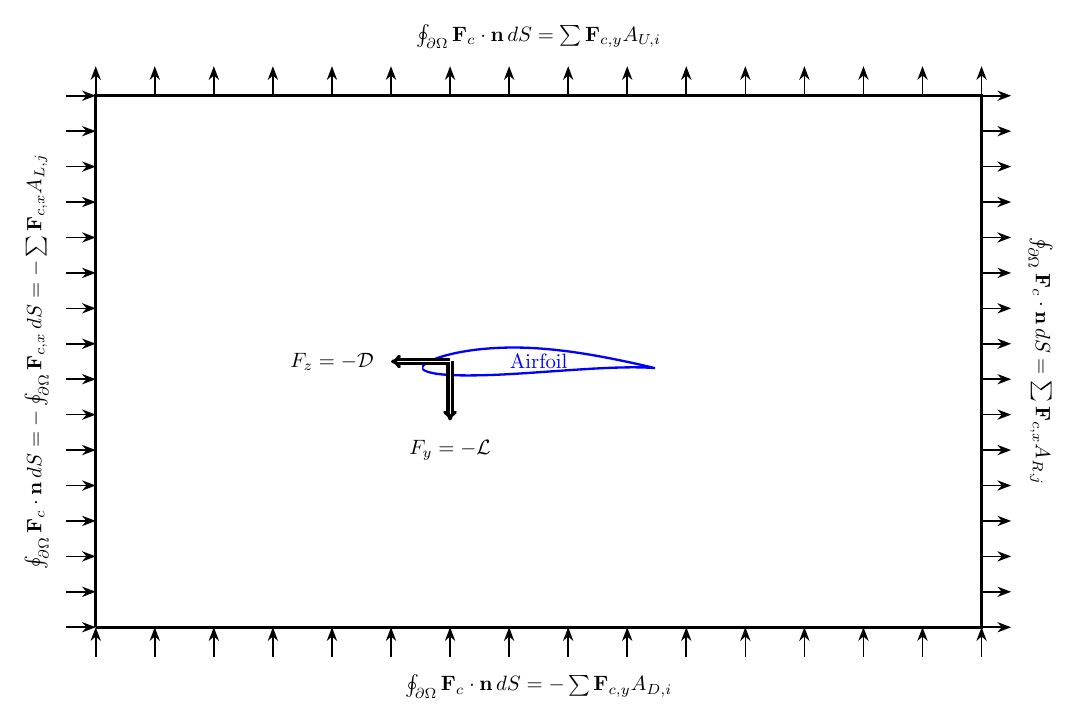}
\caption{\label{fig:Rectangular-control-volume}Rectangular control volume around an airfoil used for force calculations}
\end{figure}

The methodology for calculating the net aerodynamic forces (Lift, $L$, and Drag, $D$) acting on the airfoil is achieved by applying the integral form of the momentum conservation equation to the large, rectangular control volume (CV) that encompasses the entire airfoil as depicted in Figure \ref{fig:Rectangular-control-volume}. 

By placing the control surface far from the airfoil, we can make two key simplifications: Zero Internal Sources: Since the airfoil boundary is contained entirely within the selected CV, the source term $\mathbf{B}_{\text{internal}}$ applied across the outer control surface is zero. Diminishing Viscous Effects: As the control volume boundary is moved further away from the airfoil, the velocity gradients at the boundary diminish, allowing us to approximate the viscous flux ($\mathbf{F}_{v}$) at the outer boundary as negligible.

Under these steady-state and far-field approximations, the momentum conservation equation (Eq. (3) of the main text) % \eqref{eq:conservation-integral-form})
simplifies to a balance between the net forces acting on the fluid and the net convective momentum flux across the control surface:
\begin{equation}
\oint_{\partial\Omega}\mathbf{F}_{c}\cdot\mathbf{n}\,dS=\mathbf{F}_{\text{Fluid on Airfoil}}
\label{eq:force-balance}
\end{equation}

Here, $\mathbf{F}_{\text{Fluid on Airfoil}}$ represents the net aerodynamic force exerted by the fluid on the solid airfoil surface. This force is often broken down into the Lift ($\mathbf{L}$) and Drag ($\mathbf{D}$) components.

Conversely, the net force acting on the fluid within the control volume ($\mathbf{F}_{\text{Net}}$) is given by the negative of the force on the airfoil, as shown in the figure:

\begin{equation}
\mathbf{F}_{\text{Net}}=-(\mathbf{L}+\mathbf{D})
\label{eq:net-force}
\end{equation}

Therefore, the integral of the convective momentum flux directly equals the negative of the net aerodynamic force on the airfoil:

\begin{equation}
\oint_{\partial\Omega}\mathbf{F}_{c}\cdot\mathbf{n}\,dS=-(\mathbf{L}+\mathbf{D})
\label{eq:lift-drag-integral}
\end{equation}

By calculating the momentum flux integral from the predicted flow field over the control surface, one can obtain the Lift and Drag forces acting on the airfoil, thereby providing a robust, conservation-law-compliant metric for model evaluation.

Since our aim is to provide more physically consistent predictions, testing the results against the lift and drag predictions seems reasonable. We have utilized calculation of drag and lift forces by the using the integral form of the conservation momentum on the large, far-field control volume. %Since we don't know how far is far enough, we have tested this approach for different control volume sizes.

% TODO: ML ile ilgili hiç bir şeyi buraya aktarmadım.
% Method, Training, Sonuçlar, Conclusion EKLENMELİ.
% Uygun yerlere koyalım.

%\subsection{Generalization Beyond HLLC}

%While this study employs the HLLC (Harten-Lax-van Leer Contact) Riemann solver for flux calculations, the CoFINN methodology is not restricted to this particular scheme. The mathematical formulation in Eq. \eqref{eq:riemann-flux} uses a general flux function $\mathcal{F}(\mathbf{U}_{L},\mathbf{U}_{R},\mathbf{n})$ that can be replaced with any appropriate Riemann solver or numerical flux scheme. Alternative choices include the Roe solver, AUSM family of schemes, HLL, or even simpler schemes like Lax-Friedrichs or Rusanov.

%The HLLC solver was chosen for this work due to its excellent accuracy and robustness across the subsonic-to-supersonic flow regime encountered in transonic airfoil applications. It provides precise resolution of contact discontinuities and shock waves while maintaining computational efficiency. However, for different physical applications or flow regimes, practitioners may select the most appropriate flux calculation method without modifying the core CoFINN training framework. This flexibility makes the approach applicable to a broader range of conservation-law-driven problems beyond aerodynamics, including shallow water equations, magnetohydrodynamics, and other hyperbolic systems.

\Head{Results}
\label{sec:results}

\SubHead{Dataset}
The CNNFoil dataset contains 204 airfoil geometries from diverse airfoil families spanning a range of thickness ratios and camber values, providing geometric diversity. 
In total, there are 6{,}324 high-fidelity Reynolds-Averaged Navier-Stokes (RANS) CFD simulations at transonic conditions: Mach number $M_\infty = 0.7$, Reynolds number $Re = 6{,}000{,}000$, and angles of attack (AOA) ranging from $-10^\circ$ to $20^\circ$ with $1^\circ$ intervals. 
The working fluid is air modeled as an ideal gas with Sutherland viscosity law, and turbulence closure is provided by the Spalart-Allmaras model~\cite{spalart1992one}. %Note that the Spalart-Allmaras model, while widely used for attached boundary layer flows, has known limitations for massively separated flows at high angles of attack; the CFD training data quality therefore degrades near stall conditions ($\alpha > 15^\circ$).

The flow field data around the airfoil calculated by CFD analysis is extracted to a grid data. For this purpose, the airfoil is rotated around the quarter chord by AOA. Then the flow data is extracted in a domain of 2 chord lengths in $x$, and one chord lengths in $y$ to a grid size of $256\times 256$. The difference in the $x$ and $y$ resolutions are applied to capture the details of flow properties above the wing. For this reason, airfoil solution images provided in this study looks thicker than they actually are, and should be assessed considering this fact. All primitive variables undergo standardization, i.e., projection to zero-mean unit-variance space (z-score normalization), during training. Predictions are projected back to physical space (denormalization) before flux calculations to ensure physically meaningful momentum fluxes.

Ten airfoils from different airfoil families (Avistar, USA-35, NPL~9510, SC1012R8, NACA~2421, NACA~M12, FX~61-184, NACA~747A415, NPL~9626, RAE~5215) are held out as a fixed test set (the \emph{eval partition}) across all angles of attack to prevent data leakage. These airfoils belong to distinct families; no other airfoils from these families appear in the training or validation sets. From the remaining airfoils, a validation set of equal size (10 airfoils) is randomly drawn, with the rest used for training.

To investigate how the breadth of training data affects CoFINN training, we train models on five angle-of-attack ranges of increasing size, summarized in Table~\ref{tab:aoa-ranges}. For each range, the same set of test airfoils is evaluated at all constituent angles, enabling assessment of both interpolation accuracy within the training range and the interaction between multi-AOA training and the CoFINN loss. %The network architecture remains identical across all configurations.

\begin{table}[h!]
\centering
\caption{Training angle-of-attack ranges used in the ablation study. Training samples = 184 airfoils $\times$ \# train AOAs. All ranges are tested on 10 held-out airfoils at each constituent AOA.}
\label{tab:aoa-ranges}
%\footnotesize
\small
\begin{tabular}{llrrrr}
\hline
Name & AOA range & \# AOAs & Train samples & Validation samples & Test cases \\
\hline
Single & $0^\circ$                                   &  1 &   184 &    10 &    10 \\
Narrow & $-4^\circ$ to $4^\circ$ ($4^\circ$ step)    &  3 &   552 &    30 &    30 \\
Medium & $-8^\circ$ to $8^\circ$ ($4^\circ$ step)    &  5 &   920 &    50 &    50 \\
Broad  & $-10^\circ$ to $18^\circ$ ($2^\circ$ step)  & 15 & 2{,}760 &   150 &   150 \\
Full   & $-10^\circ$ to $20^\circ$ ($1^\circ$ step)  & 31 & 5{,}704 &   310 &   310 \\
\hline
\end{tabular}
\end{table}

The progression from Single to Full allows us to disentangle two effects: how additional training angles improve generalization across the AOA envelope, and whether the CoFINN loss remains beneficial as the data-driven baseline strengthens with more training samples. 
%Results for each range are reported in Section~\ref{sec:results}.

%%%%%%%%%%%%%%% NETWORK DETAILS %%%%%%%%%%%%%%%%%%%%%%

\SubHead{Network Architectures}

To demonstrate that the CoFINN training methodology is architecture-agnostic, we evaluate four fundamentally different network families. All models share the same input representation---a $256 \times 256$ single-channel distance field (DF)---and predict four primitive flow variables ($p$, $u$, $v$, $T$) at the same resolution. Table~\ref{tab:architectures} summarizes the key characteristics.
\begin{table}[h]
\caption{Network architectures evaluated with the CoFINN training methodology.}
\centering
\label{tab:architectures}
\renewcommand{\arraystretch}{1.1}
%\footnotesize
\small
\begin{tabular}{@{}llrrr@{}}
\toprule
Model & Type & Params & Depth & Passes \\
\midrule
CNNFoil~\cite{Duru2020} & Encoder-decoder CNN & 4.8M & 14 & 1 \\
FNO~\cite{li2021fourier} & Fourier Neural Operator & 2M & 4 & 1 \\
ViT~\cite{dosovitskiy2021image} & Vision Transformer & 5.5M & 6 & 1 \\
Diffusion~\cite{bastek_physics-informed_2024} & Conditional denoising U-Net & 8.9M & 18 & 4/20 \\
\bottomrule
\end{tabular}
\end{table}
\paragraph{CNNFoil} is the baseline encoder-decoder architecture from~\cite{Duru2020,duru_deep_2022}, consisting of four independent sub-networks that each predict one flow variable. The encoder compresses spatial dimensions from $256{\times}256$ to $2{\times}2$ through strided convolutions, and the decoder reconstructs the full-resolution field via transposed convolutions.

\paragraph{FNO} (Fourier Neural Operator)~\cite{li2021fourier} learns mappings between function spaces through spectral convolutions in the Fourier domain. Four FNO blocks with 12 Fourier modes per dimension provide a global receptive field, enabling the network to capture long-range flow interactions without pooling operations.

\paragraph{ViT} (Vision Transformer)~\cite{dosovitskiy2021image} divides the input DF into $16{\times}16$ non-overlapping patches, projects them into a 384-dimensional embedding space, and processes the resulting sequence through 6 transformer blocks with 6-head self-attention. A decoder head reconstructs the spatial output from the patch representations. While Zuo et al.~\cite{zuo2023prediction} applied a transformer encoder to airfoil flow prediction, their architecture fuses patch features with point coordinates and wall distance via an MLP, making it a point-query model rather than a purely image-to-image mapping. We adopt the standard ViT to maintain a consistent image-to-image interface across all architectures.

\paragraph{Diffusion} is a conditional denoising diffusion probabilistic model (DDPM) using the U-Net architecture from~\cite{bastek_physics-informed_2024}. The backbone follows a 4-level U-Net with channel multipliers $(1,2,4,8)$, ResNet blocks with time-conditioned adaptive group normalization, spatial linear attention at every level, and full self-attention at the bottleneck. A cosine noise schedule with 100 diffusion steps is used. The model is trained as a DDPM (denoising diffusion probabilistic model) but sampled with the deterministic DDIM (denoising diffusion implicit model) scheme for efficiency: 4 DDIM steps during training and 20 at inference. The DF input conditions the U-Net via a gradient embedding module concatenated to the initial features.

\SubHead{Training Strategy}

All models are trained for 400 epochs with the combined loss (Eq. \eqref{eq:total-loss}) active throughout. The base optimizer is Adam ($\beta_1{=}0.9$, $\beta_2{=}0.999$, $\epsilon{=}10^{-8}$) with an initial learning rate of $3 \times 10^{-3}$. The weight $\lambda$ is treated as a hyperparameter and systematically varied across $\{0, 0.1, 0.3, 0.5, 0.7, 1.0\}$ in the ablation study; $\lambda=0$ reduces the model to a purely data-driven mean absolute error (MAE) baseline, while $\lambda=1.0$ (pure CoFINN loss) caused training divergence in all configurations.
To disentangle the effect of the CoFINN from optimization improvements, we evaluate several training configurations combining standard optimization and regularization techniques (learning-rate scheduling, EMA, SWA, lookahead, early stopping), summarized in Table~\ref{tab:training-configs}. 
Each configuration is trained across the $\lambda$ values above and 3 random seeds; per (model, range, $\lambda$) cell we report the best configuration.

\begin{table}[h]
\centering
\caption{Training configurations used in the ablation study.}
\label{tab:training-configs}
\small
\begin{tabular}{ll}
\hline
Configuration & Description \\
\hline
Baseline       & Adam optimizer, fixed learning rate \\
Scheduler      & Cosine annealing (20 warmup epochs, min LR $10^{-7}$) \\
EMA            & Exponential moving average of weights (decay$=0.999$) \\
Sched+EMA      & Cosine scheduler + EMA \\
Lookahead (LA) & Lookahead~\cite{zhang_lookahead_2019} wrapper ($k{=}5$, $\alpha{=}0.5$) \\
Sched+LA       & Cosine scheduler + Lookahead \\
Sched+EStop    & Cosine scheduler + early stopping (patience$=40$) \\
Aug (v-flip)   & Vertical flip (reflects about chord line) \\
Aug (h-flip)   & Horizontal flip (reverses flow direction) \\
Aug (both)     & Combined horizontal and vertical flip \\
\hline
\end{tabular}
\end{table}

\SubHead{Evaluation Metrics}
The CoFINN error defined in \eqref{eq:pixel-flux-balance}  
is a vector, since $\textbf{F}$  is a vector. Therefore, it is possible to involve each flux loss in our calculations. We have targeted to a better prediction of forces, therefore we decided to use a momentum-based evaluation for our tests. The force estimations are evaluated with the Drag ($C_D$) and Lift $C_L$ coefficients. We employ the force-calculation method described in Supplementary Text. %Section~\ref{supp-sec:force-calc}.

We report both relative and absolute force coefficient errors. The relative errors are:
\begin{equation}
\epsilon_{C_D} = \frac{|C_{D,\text{CFD}} - C_{D,\text{pred}}|}{|C_{D,\text{CFD}}|} \times 100\%, \qquad
\epsilon_{C_L} = \frac{|C_{L,\text{CFD}} - C_{L,\text{pred}}|}{|C_{L,\text{CFD}}|} \times 100\%
\label{eq:relative-errors}
\end{equation}
The absolute errors are:
\begin{equation}
|\Delta C_D| = |C_{D,\text{CFD}} - C_{D,\text{pred}}|, \qquad
|\Delta C_L| = |C_{L,\text{CFD}} - C_{L,\text{pred}}|
\label{eq:absolute-errors}
\end{equation}
Since $C_L$ passes through zero near $\alpha=0^\circ$, $\epsilon_{C_L}$ becomes unbounded despite small absolute discrepancies. We report both metrics and note where $\epsilon_{C_L}$ is dominated by near-zero denominators. Field accuracy is measured by the normalized MAE on standardized fields (Eq.~\eqref{eq:mae-loss}), and conservation compliance by $\mathcal{L}_{\text{flux}}$ (Eq.~\eqref{eq:flux-loss}).

All results are reported as the best across 6 training improvements (schedule/EMA/stochastic weight averaging (SWA)/lookahead combinations) $\times$ 3 random seeds, selected independently per (model, AOA range, loss) cell. The hyperparameter $\lambda$ scales the rectangular control-volume force-supervision term built on Eq.~S17 %\eqref{supp-eq:lift-drag-integral} 
on supplementary document; $\lambda=0$ denotes MAE-only baseline. The per-configuration ablation is in Supplementary~Table~S6. %\ref{supp-tab:ablation-lambda}.

\SubHead{Effect of Physics Loss Weight $\lambda$}

Table~\ref{tab:lambda-sweep} shows how force prediction accuracy varies with $\lambda$ at the Full AOA range (31 evaluation angles), reporting the best configuration (improvement $\times$ seed) per $(\text{model}, \lambda)$ cell.

\begin{table}[ht]
\caption{Drag and lift coefficient error (\% mean$\pm$std across 3 seeds) vs.\ $\lambda$, eval partition, Full AOA range. Each cell picks the best improvement (of 6 for CNNFoil/FNO/ViT, 3 for Diffusion) by mean and reports that improvement's seed-wise std (averaged over 31 angles $\times$ 10 test airfoils, $|C_{D,\text{true}}|{>}0.01$, $|C_{L,\text{true}}|{>}0.05$ filters). Best $\lambda$ per model in bold. Validation-partition counterpart is Table~S1 %\ref{supp-tab:lambda-sweep-val} 
in Supplementary Information.}
\centering
\label{tab:lambda-sweep}
\renewcommand{\arraystretch}{1.1}
%\footnotesize
\small
\begin{tabular}{@{}llcc@{}}
\toprule
Model & $\lambda$ & $\langle\epsilon_{C_D}\rangle$ (\%) & $\langle\epsilon_{C_L}\rangle$ (\%) \\
\midrule
CNNFoil & $0$   & 53.8\tiny$\pm$4.7 & 22.6\tiny$\pm$0.3 \\
CNNFoil & $0.1$ & \textbf{14.8\tiny$\pm$0.8} & 20.0\tiny$\pm$0.5 \\
CNNFoil & $0.3$ & 16.4\tiny$\pm$1.8 & 20.1\tiny$\pm$0.7 \\
CNNFoil & $0.5$ & 17.8\tiny$\pm$6.4 & 19.6\tiny$\pm$0.4 \\
CNNFoil & $0.7$ & 19.5\tiny$\pm$1.6 & \textbf{19.0\tiny$\pm$1.0} \\
\midrule
FNO & $0$   & 31.7\tiny$\pm$0.2 & \textbf{19.0\tiny$\pm$0.2} \\
FNO & $0.1$ & \textbf{13.6\tiny$\pm$1.4} & 22.8\tiny$\pm$0.1 \\
FNO & $0.3$ & 13.7\tiny$\pm$1.8 & 23.0\tiny$\pm$0.7 \\
FNO & $0.5$ & 14.3\tiny$\pm$2.1 & 22.8\tiny$\pm$1.0 \\
FNO & $0.7$ & 15.5\tiny$\pm$2.8 & 22.5\tiny$\pm$0.3 \\
\midrule
ViT & $0$   & 29.0\tiny$\pm$0.6 & \textbf{17.7\tiny$\pm$0.2} \\
ViT & $0.1$ & 17.6\tiny$\pm$1.9 & 20.5\tiny$\pm$0.6 \\
ViT & $0.3$ & \textbf{16.8\tiny$\pm$0.8} & 20.6\tiny$\pm$1.2 \\
ViT & $0.5$ & 17.6\tiny$\pm$1.4 & 21.0\tiny$\pm$0.2 \\
ViT & $0.7$ & 18.1\tiny$\pm$4.0 & 19.7\tiny$\pm$0.7 \\
\midrule
Diffusion & $0$   & \textbf{52.6\tiny$\pm$1.9} & \textbf{49.0\tiny$\pm$2.8} \\
Diffusion & $0.1$ & 77.7\tiny$\pm$7.5 & 64.0\tiny$\pm$4.9 \\
Diffusion & $0.3$ & 87.8\tiny$\pm$24.0 & 58.1\tiny$\pm$4.4 \\
Diffusion & $0.5$ & 80.1\tiny$\pm$12.1 & 61.0\tiny$\pm$4.1 \\
Diffusion & $0.7$ & 101.6\tiny$\pm$14.4 & 57.4\tiny$\pm$5.0 \\
\bottomrule
\end{tabular}
\end{table}

CNNFoil achieves the largest reduction in drag coefficient error, $\epsilon_{C_D}$, decreasing from $53.8\%$ to $14.8\%$ at $\lambda = 0.1$, corresponding to a $3.6\times$ improvement. Similarly, FNO reduces $\epsilon_{C_D}$ from $31.7\%$ to $13.6\%$ at $\lambda = 0.1$ ($2.3\times$ improvement), while ViT decreases it from $29.0\%$ to $16.8\%$ at $\lambda = 0.3$ ($1.7\times$ improvement).
For the lift coefficient error, $\epsilon_{C_L}$, CNNFoil shows a modest improvement, reaching $19.0\%$ at $\lambda = 0.7$ compared to the $22.6\%$ baseline. In contrast, both FNO and ViT exhibit degraded performance on the Full dataset: FNO increases from $19.0\%$ to $22.5\%$ at $\lambda = 0.7$, while ViT increases from $17.7\%$ to $19.7\%$ at the same $\lambda$ value.
For the diffusion model trained on the Full dataset, the lowest $\epsilon_{C_D}$ is obtained at the baseline configuration ($52.6\%$), whereas all $\lambda > 0$ settings yield substantially higher errors in the range of $77$--$102\%$. Likewise, the minimum $\epsilon_{C_L}$ is also achieved at the baseline ($49.0\%$). However, under narrower training ranges (Table~\ref{tab:range-improvement}), CoFINN consistently reduces both $\epsilon_{C_D}$ and $\epsilon_{C_L}$ for the diffusion model.

%CNNFoil drops drag coefficient error, $\epsilon_{C_D}$ from $53.8\%$ to $14.8\%$ at $\lambda{=}0.1$ ($3.6\times$ improvement); FNO from $31.7\%$ to $13.6\%$ at $\lambda{=}0.1$ ($2.3\times$); ViT from $29.0\%$ to $16.8\%$ at $\lambda{=}0.3$ ($1.7\times$). 

%For lift coefficient, CNNFoil improves to $19.0\%$ at $\lambda{=}0.7$ (vs $22.6\%$ baseline); FNO regresses on Full ($19.0\%\to22.5\%$ at $\lambda{=}0.7$); ViT regresses on Full ($17.7\%\to19.7\%$ at $\lambda{=}0.7$). 

%Diffusion at Full has its lowest $\epsilon_{C_D}$ at the baseline ($52.6\%$, all $\lambda{>}0$ values $77$--$102\%$) and lowest $\epsilon_{C_L}$ at the baseline ($49.0\%$). At narrower training ranges (Table~\ref{tab:range-improvement}), CoFINN reduces both $\epsilon_{C_D}$ and $\epsilon_{C_L}$ for diffusion model.

\SubHead{Improvement Across Training Data Regimes}

Table~\ref{tab:range-improvement} reports the best-of-best configuration per $(\text{model}, \text{range}, \text{loss})$ cell: each cell independently selects the (improvement, seed, $\lambda$) with lowest drag error under the indicated loss ($\lambda=0$ for MAE-only, $\lambda>0$ for CoFINN). Values are mean absolute percentage error on the 10 test airfoils, averaged over the evaluation angles of the training range.

\begin{table}[ht]
\centering
\caption{Drag and lift coefficient error (\% mean$\pm$std across 3 seeds) by training AOA range, MAE-only ($\lambda{=}0$) vs.\ CoFINN (best $\lambda{>}0$). Each cell picks the best (improvement) by mean over seeds and reports that improvement's seed-wise std (averaged over training-range AOAs and 10 test airfoils, with $|C_{D,\text{true}}|{>}0.01$ and $|C_{L,\text{true}}|{>}0.05$ to avoid divergence from near-zero denominators). Parenthesized ($\lambda$) is the winning weight for CoFINN. $\Delta$: relative reduction in mean ($+$ is better).}
\label{tab:range-improvement}
\renewcommand{\arraystretch}{1.1}
%\footnotesize
\small
\setlength{\tabcolsep}{4pt}
\begin{tabular}{@{}llccccccc@{}}
\toprule
& & \multicolumn{3}{c}{$\epsilon_{C_D}$ (\%)} & \multicolumn{3}{c}{$\epsilon_{C_L}$ (\%)} \\
\cmidrule(lr){3-5} \cmidrule(lr){6-8}
Model & Range & MAE-only & CoFINN ($\lambda$) & $\Delta_{C_D}$ & MAE-only & CoFINN ($\lambda$) & $\Delta_{C_L}$ \\
\midrule
CNNFoil   & Single & 95.0\tiny$\pm$69.8 & \textbf{13.8\tiny$\pm$4.4} (0.3) & \textbf{+85\%} & 31.5\tiny$\pm$5.7 & \textbf{15.2\tiny$\pm$3.8} (0.5) & \textbf{+52\%} \\
CNNFoil   & Narrow & 107.7\tiny$\pm$15.1 & \textbf{18.9\tiny$\pm$5.4} (0.7) & \textbf{+82\%} & 34.7\tiny$\pm$2.7 & \textbf{23.2\tiny$\pm$2.0} (0.1) & \textbf{+33\%} \\
CNNFoil   & Medium & 82.8\tiny$\pm$15.2 & \textbf{18.8\tiny$\pm$6.2} (0.3) & \textbf{+77\%} & 30.3\tiny$\pm$0.9 & \textbf{19.5\tiny$\pm$1.8} (0.3) & \textbf{+36\%} \\
CNNFoil   & Broad  & 53.5\tiny$\pm$2.3 & \textbf{15.3\tiny$\pm$1.3} (0.1) & \textbf{+71\%} & 21.5\tiny$\pm$0.3 & \textbf{18.4\tiny$\pm$0.5} (0.7) & \textbf{+14\%} \\
CNNFoil   & Full   & 53.8\tiny$\pm$4.7 & \textbf{14.8\tiny$\pm$0.8} (0.1) & \textbf{+73\%} & 22.6\tiny$\pm$0.3 & \textbf{19.0\tiny$\pm$1.0} (0.7) & \textbf{+16\%} \\
\midrule
FNO       & Single & 33.6\tiny$\pm$16.1 & \textbf{12.5\tiny$\pm$0.7} (0.5) & \textbf{+63\%} & 16.3\tiny$\pm$1.2 & \textbf{12.9\tiny$\pm$4.1} (0.5) & \textbf{+21\%} \\
FNO       & Narrow & 37.6\tiny$\pm$4.4 & \textbf{22.5\tiny$\pm$4.4} (0.5) & \textbf{+40\%} & \textbf{18.4\tiny$\pm$0.5} & 23.6\tiny$\pm$1.2 (0.7) & $-$28\% \\
FNO       & Medium & 26.9\tiny$\pm$3.9 & \textbf{18.5\tiny$\pm$1.0} (0.7) & \textbf{+31\%} & \textbf{18.0\tiny$\pm$0.7} & 26.0\tiny$\pm$0.7 (0.1) & $-$44\% \\
FNO       & Broad  & 31.7\tiny$\pm$0.6 & \textbf{12.3\tiny$\pm$0.4} (0.1) & \textbf{+61\%} & \textbf{17.1\tiny$\pm$0.2} & 20.5\tiny$\pm$0.6 (0.7) & $-$20\% \\
FNO       & Full   & 31.7\tiny$\pm$0.2 & \textbf{13.6\tiny$\pm$1.4} (0.1) & \textbf{+57\%} & \textbf{19.0\tiny$\pm$0.2} & 22.5\tiny$\pm$0.3 (0.7) & $-$19\% \\
\midrule
ViT       & Single & 24.6\tiny$\pm$16.7 & \textbf{12.1\tiny$\pm$4.6} (0.7) & \textbf{+51\%} & \textbf{13.1\tiny$\pm$1.4} & 18.8\tiny$\pm$8.3 (0.7) & $-$44\% \\
ViT       & Narrow & 47.4\tiny$\pm$6.7 & \textbf{31.2\tiny$\pm$3.0} (0.3) & \textbf{+34\%} & 59.8\tiny$\pm$3.0 & \textbf{38.8\tiny$\pm$7.9} (0.5) & \textbf{+35\%} \\
ViT       & Medium & 35.2\tiny$\pm$5.2 & \textbf{25.9\tiny$\pm$3.1} (0.7) & \textbf{+26\%} & 35.8\tiny$\pm$3.6 & \textbf{35.3\tiny$\pm$3.4} (0.1) & +1\% \\
ViT       & Broad  & 27.1\tiny$\pm$0.4 & \textbf{16.6\tiny$\pm$1.2} (0.3) & \textbf{+39\%} & \textbf{16.1\tiny$\pm$0.3} & 19.3\tiny$\pm$1.2 (0.1) & $-$20\% \\
ViT       & Full   & 29.0\tiny$\pm$0.6 & \textbf{16.8\tiny$\pm$0.8} (0.3) & \textbf{+42\%} & \textbf{17.7\tiny$\pm$0.2} & 19.7\tiny$\pm$0.7 (0.7) & $-$12\% \\
\midrule
Diffusion & Single & 199.3\tiny$\pm$28.3 & \textbf{78.8\tiny$\pm$41.8} (0.1) & \textbf{+60\%} & 59.9\tiny$\pm$10.4 & \textbf{48.1\tiny$\pm$1.6} (0.3) & \textbf{+20\%} \\
Diffusion & Narrow & 140.2\tiny$\pm$27.5 & \textbf{109.1\tiny$\pm$18.7} (0.1) & \textbf{+22\%} & 87.3\tiny$\pm$4.2 & \textbf{73.7\tiny$\pm$16.4} (0.3) & \textbf{+16\%} \\
Diffusion & Medium & 133.1\tiny$\pm$9.0 & \textbf{87.7\tiny$\pm$31.5} (0.5) & \textbf{+34\%} & 93.0\tiny$\pm$1.7 & \textbf{78.4\tiny$\pm$10.5} (0.1) & \textbf{+16\%} \\
Diffusion & Broad  & \textbf{65.5\tiny$\pm$4.8} & 81.3\tiny$\pm$6.3 (0.1) & $-$24\% & \textbf{54.3\tiny$\pm$6.2} & 57.3\tiny$\pm$4.0 (0.3) & $-$6\% \\
Diffusion & Full & \textbf{52.6\tiny$\pm$1.9} & 77.7\tiny$\pm$7.5 (0.1) & $-$48\% & \textbf{49.0\tiny$\pm$2.8} & 57.4\tiny$\pm$5.0 (0.7) & $-$17\% \\
\bottomrule
\end{tabular}
\vspace{2pt}

%{\footnotesize $^*$Diffusion Full results are provisional---training still in progress as of writing; values from latest checkpoint snapshot.}
\end{table}

CNNFoil achieves the most substantial drag coefficient error reduction across all five training ranges, with $\epsilon_{C_D}$ decreasing by approximately $71$--$85\%$: from $95.0\%$ to $13.8\%$ on Single, $107.7\%$ to $18.9\%$ on Narrow, $82.8\%$ to $18.8\%$ on Medium, $53.5\%$ to $15.3\%$ on Broad, and $53.8\%$ to $14.8\%$ on Full. The lift coefficient error, $\epsilon_{C_L}$, is also consistently reduced, with improvements ranging from $14$--$52\%$ across the same training regimes.
FNO likewise shows consistent improvements in drag prediction accuracy, reducing $\epsilon_{C_D}$ by approximately $31$--$63\%$: $33.6\% \to 12.5\%$ (Single), $37.6\% \to 22.5\%$ (Narrow), $26.9\% \to 18.5\%$ (Medium), $31.7\% \to 12.3\%$ (Broad), and $31.7\% \to 13.6\%$ (Full). However, the behavior for lift prediction is mixed. While $\epsilon_{C_L}$ improves by $21\%$ on the Single range, it deteriorates by approximately $19$--$44\%$ for the Narrow, Medium, Broad, and Full ranges.
ViT reduces drag coefficient error by approximately $26$--$51\%$ across all ranges, with $\epsilon_{C_D}$ decreasing from $24.6\%$ to $12.1\%$ (Single), $47.4\%$ to $31.2\%$ (Narrow), $35.2\%$ to $25.9\%$ (Medium), $27.1\%$ to $16.6\%$ (Broad), and $29.0\%$ to $16.8\%$ (Full). For lift prediction, ViT improves performance on the Narrow and Medium ranges ($59.8\% \to 38.8\%$ and $35.8\% \to 35.3\%$, respectively), but exhibits degradation of approximately $12$--$44\%$ on the Single, Broad, and Full configurations.
The diffusion model demonstrates more variable behavior. Drag coefficient error decreases by approximately $22$--$60\%$ for the Single, Narrow, and Medium ranges ($199.3\% \to 78.8\%$, $140.2\% \to 109.1\%$, and $133.1\% \to 87.7\%$, respectively), but worsens by approximately $24$--$48\%$ on the Broad and Full ranges ($65.5\% \to 81.3\%$ and $52.6\% \to 77.7\%$). A similar trend is observed for lift prediction: $\epsilon_{C_L}$ improves by approximately $16$--$20\%$ for Single, Narrow, and Medium, while degrading by approximately $6$--$17\%$ for Broad and Full.

%CNNFoil drag drops by $71$--$85\%$ across all five ranges ($95.0\% \to 13.8\%$ Single; $107.7\% \to 18.9\%$ Narrow; $82.8\% \to 18.8\%$ Medium; $53.5\% \to 15.3\%$ Broad; $53.8\% \to 14.8\%$ Full). Lift drops by $14$--$52\%$ on the same ranges.

%FNO drag drops by $31$--$63\%$ ($33.6\% \to 12.5\%$ Single; $37.6\% \to 22.5\%$ Narrow; $26.9\% \to 18.5\%$ Medium; $31.7\% \to 12.3\%$ Broad; $31.7\% \to 13.6\%$ Full). Lift improves at Single ($+21\%$) and regresses by $19$--$44\%$ on Narrow/Medium/Broad/Full.

%ViT drag drops by $26$--$51\%$ ($24.6\% \to 12.1\%$ Single; $47.4\% \to 31.2\%$ Narrow; $35.2\% \to 25.9\%$ Medium; $27.1\% \to 16.6\%$ Broad; $29.0\% \to 16.8\%$ Full). Lift drops at Narrow/Medium ($59.8\% \to 38.8\%$ Narrow; $35.8\% \to 35.3\%$ Medium) and regresses by $12$--$44\%$ on Single/Broad/Full.

%Diffusion drag drops by $22$--$60\%$ on Single/Narrow/Medium ($199.3\% \to 78.8\%$ Single; $140.2\% \to 109.1\%$ Narrow; $133.1\% \to 87.7\%$ Medium) and regresses by $24$--$48\%$ on Broad/Full ($65.5\% \to 81.3\%$ Broad; $52.6\% \to 77.7\%$ Full). Lift follows the same pattern: $+16$--$20\%$ on Single/Narrow/Medium, $-6$--$17\%$ on Broad/Full. Full-range diffusion values are provisional pending completion of all training configurations.

\SubHead{Per-Angle Force Error Analysis}

Tables~\ref{tab:per-aoa-cd} (and~S2 %\ref{supp-tab:per-aoa-cl} 
in supplementary text) break down drag and lift errors at six representative angles for the full training range.

\begin{table}[ht]
\centering
\caption{Drag coefficient error $\epsilon_{C_D}$ (\% mean$\pm$std across 3 seeds, eval partition, Full range = 31 evaluation angles). Each cell picks the best improvement (of 6) by mean and reports that improvement's seed-wise std (with $|C_{D,\text{true}}|{>}0.01$ filter). $\langle\text{all}\rangle$ is the best improvement's per-AOA mean averaged over all 31 angles. Best $\lambda$ per column in bold.}
\label{tab:per-aoa-cd}
\renewcommand{\arraystretch}{1.1}
%\footnotesize
\small
\setlength{\tabcolsep}{2pt}
\begin{tabular}{@{}llccccccc@{}}
\toprule
Model & $\lambda$ & $-10^\circ$ & $-4^\circ$ & $0^\circ$ & $8^\circ$ & $12^\circ$ & $20^\circ$ & $\langle\text{all}\rangle$ \\
\midrule
CNNFoil & $0$   & 36.4\tiny$\pm$4.9 & 93.9\tiny$\pm$6.9 & 100.5\tiny$\pm$24.1 & 18.0\tiny$\pm$3.1 & 24.5\tiny$\pm$2.2 & 73.4\tiny$\pm$2.1 & 57.0 \\
CNNFoil & $0.1$ & 20.4\tiny$\pm$1.0 & 15.3 & 25.4\tiny$\pm$3.5 & \textbf{3.3} & 5.0\tiny$\pm$0.3 & 22.5\tiny$\pm$0.6 & \textbf{15.5} \\
CNNFoil & $0.3$ & \textbf{17.2} & 23.9\tiny$\pm$2.0 & \textbf{19.8\tiny$\pm$10.3} & 7.1 & \textbf{3.8\tiny$\pm$1.2} & 17.4\tiny$\pm$1.4 & 15.8 \\
CNNFoil & $0.5$ & 18.8\tiny$\pm$2.1 & \textbf{9.1} & 47.3\tiny$\pm$30.6 & 7.8\tiny$\pm$0.3 & 3.9 & \textbf{15.1\tiny$\pm$8.1} & 19.4 \\
CNNFoil & $0.7$ & 21.0\tiny$\pm$1.5 & 28.5 & 21.7 & 8.2\tiny$\pm$0.2 & 4.9\tiny$\pm$1.2 & 29.2\tiny$\pm$4.2 & 21.1 \\
\midrule
FNO & $0$   & 27.8\tiny$\pm$5.8 & 28.4\tiny$\pm$6.6 & 30.8\tiny$\pm$8.8 & 9.7\tiny$\pm$0.4 & 22.0\tiny$\pm$3.0 & 72.3\tiny$\pm$3.2 & 32.1 \\
FNO & $0.1$ & \textbf{11.9\tiny$\pm$2.0} & 22.4\tiny$\pm$3.0 & \textbf{14.4\tiny$\pm$3.7} & \textbf{5.9\tiny$\pm$0.5} & 7.4\tiny$\pm$1.1 & 15.6\tiny$\pm$9.7 & \textbf{14.1} \\
FNO & $0.3$ & 15.4\tiny$\pm$0.8 & \textbf{18.1\tiny$\pm$1.5} & 21.1\tiny$\pm$6.9 & 7.3\tiny$\pm$0.9 & 7.1\tiny$\pm$0.6 & \textbf{15.1\tiny$\pm$6.0} & 14.4 \\
FNO & $0.5$ & 17.5\tiny$\pm$0.9 & 18.6\tiny$\pm$4.6 & 25.9\tiny$\pm$4.9 & 7.5\tiny$\pm$1.1 & 7.0\tiny$\pm$1.0 & 20.1\tiny$\pm$13.2 & 15.3 \\
FNO & $0.7$ & 16.9\tiny$\pm$3.1 & 23.6\tiny$\pm$10.1 & 22.6\tiny$\pm$3.6 & 7.7\tiny$\pm$2.0 & \textbf{5.8\tiny$\pm$0.3} & 27.5\tiny$\pm$14.5 & 16.7 \\
\midrule
ViT & $0$   & 20.3\tiny$\pm$6.4 & \textbf{14.7\tiny$\pm$2.4} & 20.3\tiny$\pm$1.4 & \textbf{7.1\tiny$\pm$0.1} & 20.8\tiny$\pm$5.5 & 73.7\tiny$\pm$3.0 & 29.2 \\
ViT & $0.1$ & \textbf{16.5\tiny$\pm$1.8} & 35.4\tiny$\pm$17.7 & \textbf{16.6\tiny$\pm$5.8} & 8.0\tiny$\pm$1.2 & \textbf{5.5\tiny$\pm$0.9} & 23.6\tiny$\pm$3.8 & 17.7 \\
ViT & $0.3$ & 18.5\tiny$\pm$0.4 & 29.3\tiny$\pm$3.3 & 17.7\tiny$\pm$2.7 & 9.0\tiny$\pm$3.2 & 6.1\tiny$\pm$0.1 & 24.8\tiny$\pm$1.5 & \textbf{17.0} \\
ViT & $0.5$ & 17.9\tiny$\pm$3.1 & 34.9\tiny$\pm$10.1 & 20.1\tiny$\pm$8.1 & 7.1\tiny$\pm$0.3 & 5.8\tiny$\pm$0.9 & \textbf{10.6\tiny$\pm$3.4} & 17.9 \\
ViT & $0.7$ & 18.9\tiny$\pm$3.9 & 28.2\tiny$\pm$7.4 & 21.5\tiny$\pm$1.4 & 8.9\tiny$\pm$1.9 & 5.6\tiny$\pm$1.0 & 17.6\tiny$\pm$8.3 & 18.3 \\
\midrule
\multicolumn{9}{l}{\textit{Diffusion Full-range results are reported in the Supplementary Information.}} \\
\bottomrule
\end{tabular}
\end{table}

CoFINN consistently drops CNNFoil's drag coefficient error, $\epsilon_{C_D}$ at every AOA (e.g.\ $-10^\circ$: $36.4\%\to 17.2\%$; $0^\circ$: $100.5\% \to 19.8\%$; $20^\circ$: $73.4\% \to 15.1\%$). The largest gains occur near $\alpha{=}0^\circ$, where $C_D$ is small and the baseline underpredicts it most severely (often to near-zero or negative). FNO exhibits a similar pattern but with a smaller starting error. ViT's per-AOA response is more uneven, reflecting its sensitivity to training range discussed above. The optimal $\lambda$ varies by AOA: larger values ($0.5$--$0.7$) perform best at negative or high-positive angles, while moderate values ($0.1$--$0.3$) achieve the lowest average error.

\SubHead{Absolute Force Coefficient Predictions}

Tables~\ref{tab:absolute-cd} (and~S4 %\ref{supp-tab:absolute-cl} 
in supplementary text) present predicted $C_D$ and $C_L$ values alongside CFD ground truth. This complements Tables~\ref{tab:per-aoa-cd} and~S2: %\ref{supp-tab:per-aoa-cl}: 
relative errors can be misleading when denominators are small, so absolute values show physical magnitudes directly. Each cell is the best-of-best run (6 improvements $\times$ 3 seeds) per (model, $\lambda$, AOA); bold marks the $\lambda$ closest to GT per column.

\begin{table}[ht]
\centering
\caption{Predicted vs.\ CFD drag coefficient $C_D$ at representative angles, eval partition, Full range, GT averaged over 10 fixed test airfoils. Best of all improvements $\times$ 3 seeds per cell, picked by closeness to GT. Bold: $\lambda$ closest to GT per column. Validation-partition counterpart is Table~S3 %\ref{supp-tab:absolute-cd-val} 
in Supplementary Information.}
\label{tab:absolute-cd}
\renewcommand{\arraystretch}{1.1}
%\footnotesize
\small
\setlength{\tabcolsep}{3pt}
\begin{tabular}{@{}llcccccc@{}}
\toprule
Model & $\lambda$ & $-10^\circ$ & $-4^\circ$ & $0^\circ$ & $8^\circ$ & $12^\circ$ & $20^\circ$ \\
\midrule
CFD (GT) & --- & 0.123 & 0.032 & 0.021 & 0.113 & 0.188 & 0.392 \\
\midrule
CNNFoil & $0$   & 0.083 & 0.002 & 0.003 & \textbf{0.107} & 0.151 & 0.127 \\
CNNFoil & $0.1$ & 0.100 & 0.024 & 0.014 & 0.103 & 0.183 & 0.478 \\
CNNFoil & $0.3$ & \textbf{0.103} & \textbf{0.032} & \textbf{0.020} & 0.105 & \textbf{0.192} & 0.457 \\
CNNFoil & $0.5$ & 0.101 & 0.024 & 0.015 & 0.104 & 0.183 & \textbf{0.366} \\
CNNFoil & $0.7$ & 0.099 & 0.029 & 0.016 & 0.103 & 0.183 & 0.487 \\
\midrule
 FNO & $0$   & 0.096 & \textbf{0.027} & 0.014 & \textbf{0.112} & 0.148 & 0.119 \\
FNO & $0.1$ & \textbf{0.110} & 0.024 & 0.018 & 0.108 & 0.175 & 0.377 \\
FNO & $0.3$ & 0.104 & 0.025 & 0.018 & 0.107 & 0.177 & \textbf{0.396} \\
FNO & $0.5$ & 0.102 & 0.024 & 0.016 & 0.105 & 0.179 & 0.434 \\
FNO & $0.7$ & 0.102 & 0.024 & \textbf{0.019} & 0.105 & \textbf{0.182} & 0.457 \\
\midrule
ViT & $0$   & \textbf{0.113} & 0.025 & 0.017 & \textbf{0.115} & 0.160 & 0.285 \\
ViT & $0.1$ & 0.107 & \textbf{0.026} & \textbf{0.018} & 0.105 & 0.183 & \textbf{0.386} \\
ViT & $0.3$ & 0.100 & 0.025 & 0.018 & 0.105 & 0.180 & 0.399 \\
ViT & $0.5$ & 0.105 & 0.025 & 0.018 & 0.106 & 0.181 & 0.383 \\
ViT & $0.7$ & 0.105 & 0.022 & 0.015 & 0.105 & \textbf{0.187} & 0.358 \\
\midrule
Diffusion & $0$   & 0.060 & 0.024 & \textbf{0.016} & \textbf{0.103} & 0.123 & 0.105 \\
Diffusion & $0.1$ & 0.049 & 0.010 & 0.016 & 0.127 & \textbf{0.144} & \textbf{0.325} \\
Diffusion & $0.3$ & 0.029 & \textbf{0.031} & 0.010 & 0.040 & 0.040 & 0.059 \\
Diffusion & $0.5$ & \textbf{0.070} & 0.009 & 0.008 & 0.054 & 0.242 & 0.278 \\
Diffusion & $0.7$ & 0.063 & 0.028 & 0.029 & 0.057 & 0.296 & 0.278 \\
\bottomrule
\end{tabular}
\end{table}

CNNFoil is the most physics-consistent model: the baseline ($\lambda{=}0$) collapses drag near zero at $\alpha{=}{-4}^\circ$ and $0^\circ$ (0.002 and 0.003 vs GT 0.032 and 0.021), while $\lambda{=}0.3$ recovers these to 0.032 and 0.020---matching GT to three decimal places at $\alpha{=}{-4}^\circ$. At $\alpha{=}20^\circ$, $\lambda{=}0.5$ achieves the closest drag prediction (0.366 vs GT 0.392, $6.7\%$ error). FNO and ViT both undershoot at $\alpha{=}20^\circ$ under the baseline (0.119 and 0.285 vs GT 0.392), with $\lambda{=}0.3$ closest for FNO (0.396) and $\lambda{=}0.1$ for ViT (0.386). For $C_L$, CNNFoil overshoots at $\alpha{=}20^\circ$ across all $\lambda$ (${\approx}1.3$--$1.4$ vs GT 0.990); FNO and ViT similarly overshoot (${\approx}1.0$--$1.4$), with ViT $\lambda{=}0.5$ achieving the closest result (1.022). Diffusion predictions remain noisier: $C_D$ at $\alpha{=}{-10}^\circ$ ranges from 0.029 to 0.070 vs GT 0.123, and $C_L$ at $\alpha{=}{-10}^\circ$ is positive for $\lambda{=}0$ (0.070) where GT is $-0.265$; $\lambda{=}0.7$ partially corrects the sign ($-0.233$). CoFINN supervision systematically pulls the predicted force coefficients toward the CFD values---correcting the baselines' tendency to under-predict or mis-sign $C_D$ and $C_L$---with the largest corrections at low-drag angles near $\alpha{=}0^\circ$, where the unconstrained models are least accurate.

\SubHead{Generalization to Unseen Angles}

To test generalization, we train on the Medium AOA range (5 even-degree angles ${-8, -4, 0, 4, 8}^\circ$) and evaluate on all 15 odd-degree angles that the model never saw during training. Tables~\ref{tab:unseen-cd} (and~S5 %\ref{supp-tab:unseen-cl} 
in supplementary text)
show drag and lift errors at six representative unseen angles and the mean across all 15. Each cell picks the best improvement (of 6) by mean and reports that improvement's seed-wise std; bold marks the best $\lambda$ per column.

\begin{table}[ht]
\centering
\caption{Drag error $\epsilon_{C_D}$ (\% mean$\pm$std across 3 seeds, eval partition, Medium range training, odd-degree test angles). Each cell picks the best improvement (of 6) by mean and reports that improvement's seed-wise std. $\langle\text{all}\rangle$ is the best improvement's per-AOA mean averaged over all 15 unseen angles. Best $\lambda$ per column in bold.}
\label{tab:unseen-cd}
\renewcommand{\arraystretch}{1.05}
%\scriptsize
\small
\setlength{\tabcolsep}{3pt}
\begin{tabular}{@{}llccccccc@{}}
\toprule
Model & $\lambda$ & $-9^\circ$ & $-3^\circ$ & $-1^\circ$ & $5^\circ$ & $9^\circ$ & $19^\circ$ & $\langle\text{all}\rangle$ \\
\midrule
CNNFoil & $0$   & 58.2\tiny$\pm$5.8 & 132.1\tiny$\pm$15.0 & 161.9\tiny$\pm$34.8 & 39.0\tiny$\pm$14.2 & 26.2\tiny$\pm$1.3 & 114.2\tiny$\pm$10.9 & 88.1 \\
CNNFoil & $0.1$ & 20.6\tiny$\pm$6.5 & 23.6\tiny$\pm$12.2 & 28.6\tiny$\pm$9.6 & 13.0\tiny$\pm$2.6 & 7.0\tiny$\pm$1.8 & 59.9\tiny$\pm$10.5 & 27.3 \\
CNNFoil & $0.3$ & 19.2\tiny$\pm$2.1 & \textbf{19.2} & \textbf{25.3\tiny$\pm$9.9} & 9.7\tiny$\pm$0.2 & 6.9\tiny$\pm$2.7 & 71.8\tiny$\pm$14.7 & 31.9 \\
CNNFoil & $0.5$ & 26.0 & 24.5\tiny$\pm$4.0 & 37.1\tiny$\pm$18.1 & \textbf{7.8} & 10.2 & 49.3 & 28.8 \\
CNNFoil & $0.7$ & \textbf{11.0} & 25.4\tiny$\pm$11.1 & 27.7 & 8.6\tiny$\pm$3.1 & \textbf{5.7\tiny$\pm$0.1} & \textbf{10.8} & \textbf{15.6} \\
\midrule
FNO & $0$   & 43.5\tiny$\pm$3.1 & 41.8\tiny$\pm$10.6 & 26.9\tiny$\pm$6.3 & 29.3\tiny$\pm$5.7 & 31.0\tiny$\pm$9.8 & 71.8\tiny$\pm$23.6 & 37.9 \\
FNO & $0.1$ & \textbf{18.0\tiny$\pm$0.8} & 40.2\tiny$\pm$22.6 & 27.1\tiny$\pm$13.1 & \textbf{11.9\tiny$\pm$2.3} & \textbf{12.9\tiny$\pm$2.9} & 74.0\tiny$\pm$1.1 & \textbf{33.1} \\
FNO & $0.3$ & 23.3\tiny$\pm$3.8 & \textbf{31.5\tiny$\pm$9.5} & 30.4\tiny$\pm$2.4 & 12.1\tiny$\pm$1.9 & 18.4\tiny$\pm$5.6 & 88.5\tiny$\pm$21.4 & 38.9 \\
FNO & $0.5$ & 28.5\tiny$\pm$7.0 & 33.9\tiny$\pm$6.6 & 36.4\tiny$\pm$1.4 & 12.4\tiny$\pm$0.6 & 17.8\tiny$\pm$4.3 & \textbf{61.3\tiny$\pm$53.5} & 37.1 \\
FNO & $0.7$ & 28.8\tiny$\pm$2.1 & 32.9\tiny$\pm$6.9 & \textbf{26.9\tiny$\pm$5.8} & 12.8\tiny$\pm$0.8 & 16.5\tiny$\pm$5.4 & 75.0\tiny$\pm$18.5 & 36.6 \\
\midrule
ViT & $0$   & 45.6\tiny$\pm$5.4 & 42.9\tiny$\pm$5.9 & \textbf{24.2\tiny$\pm$11.6} & \textbf{10.1\tiny$\pm$1.3} & 22.9\tiny$\pm$5.8 & 104.5\tiny$\pm$6.5 & 57.7 \\
ViT & $0.1$ & 32.3\tiny$\pm$7.8 & 29.9\tiny$\pm$6.6 & 30.3\tiny$\pm$11.1 & 18.9\tiny$\pm$8.3 & 10.5\tiny$\pm$2.1 & 54.4\tiny$\pm$28.2 & 33.2 \\
ViT & $0.3$ & 32.4\tiny$\pm$8.5 & 31.1\tiny$\pm$13.8 & 38.7\tiny$\pm$12.9 & 14.3\tiny$\pm$1.8 & 8.7\tiny$\pm$2.7 & 24.2 & 34.4 \\
ViT & $0.5$ & 37.9\tiny$\pm$6.3 & 35.4\tiny$\pm$9.1 & 33.8\tiny$\pm$2.2 & 20.4\tiny$\pm$1.9 & 11.9\tiny$\pm$1.8 & \textbf{18.1\tiny$\pm$1.6} & 36.1 \\
ViT & $0.7$ & \textbf{27.8\tiny$\pm$12.2} & \textbf{23.0\tiny$\pm$0.6} & 25.3\tiny$\pm$8.8 & 19.7\tiny$\pm$5.8 & \textbf{7.9\tiny$\pm$6.0} & 33.2\tiny$\pm$3.5 & \textbf{29.0} \\
\bottomrule
\end{tabular}
\end{table}

CoFINN substantially improves drag generalization on unseen angles across all three architectures. CNNFoil's best drag result at $\lambda{=}0.7$ ($15.6\%$ mean) is $5.6\times$ better than the baseline ($88.1\%$), with $\epsilon_{C_D}$ reduced from $114.2\%$ to $10.8\%$ at $\alpha{=}19^\circ$ and from $58.2\%$ to $11.0\%$ at $\alpha{=}{-9}^\circ$. FNO improves from $37.9\%$ (baseline) to $33.1\%$ at $\lambda{=}0.1$, and ViT from $57.7\%$ to $29.0\%$ at $\lambda{=}0.7$. Lift generalization is mixed: CNNFoil benefits from $\lambda{=}0.1$ ($29.7\%$ vs $34.5\%$ baseline), while FNO retains the baseline's advantage ($24.3\%$) and ViT shows a narrow win for $\lambda{=}0.7$ ($40.6\%$, tied with baseline but with a much flatter per-angle profile). The gains are largest at boundary angles ($\alpha{=}{-9}^\circ$, $19^\circ$) where extrapolation beyond the Medium training envelope is most severe. The full breakdown across all 15 unseen angles is in Supplementary text Tables~S10 and S11. %\ref{supp-tab:full-unseen-cd} and~\ref{supp-tab:full-unseen-cl}.

\SubHead{Generalization to Held-Out Samples}

To verify that improvement extends beyond the fixed test airfoils, we evaluate on the validation split: 10\% of training samples randomly held out as (airfoil, angle) pairs. Unlike the test set (10 fixed airfoil families at every AOA), val samples are drawn from the training airfoil pool and are not uniformly distributed across angles---the same airfoil may appear at one angle in training and a different angle in validation. Table~\ref{tab:val-summary} compares the best baseline ($\lambda{=}0$) against the best CoFINN ($\lambda{>}0$) on val samples, with each entry picked as the best-of-best run (6 improvements $\times$ 3 seeds) averaged across all val AOAs for that range.

\begin{table}[ht]
\centering
\caption{Val-split force prediction errors (\% mean$\pm$std across 3 seeds, averaged over training-range AOAs and val airfoils with $|C_{D,\text{true}}|{>}0.01$ and $|C_{L,\text{true}}|{>}0.05$ to avoid divergence from near-zero denominators). Best of all improvements $\times$ 3 seeds per cell (6 improvements for CNNFoil/FNO/ViT, 3 for Diffusion), selected by mean. Parenthesized number for $\lambda{>}0$ rows is the winning $\lambda$ value ($C_D$ and $C_L$ selected independently). Bold: better of the two loss rows per column.}
\label{tab:val-summary}
\renewcommand{\arraystretch}{1.1}
%\footnotesize
\small
\setlength{\tabcolsep}{3pt}
\begin{tabular}{@{}llcccccc@{}}
\toprule
& & \multicolumn{2}{c}{Narrow} & \multicolumn{2}{c}{Medium} & \multicolumn{2}{c}{Broad} \\
\cmidrule(lr){3-4} \cmidrule(lr){5-6} \cmidrule(lr){7-8}
Model & Loss & $\epsilon_{C_D}$ & $\epsilon_{C_L}$ & $\epsilon_{C_D}$ & $\epsilon_{C_L}$ & $\epsilon_{C_D}$ & $\epsilon_{C_L}$ \\
\midrule
\multirow{2}{*}{CNNFoil} & $\lambda{=}0$        & 57.3\tiny$\pm$8.8 & 22.7\tiny$\pm$4.0 & 40.6\tiny$\pm$5.5 & 18.8\tiny$\pm$2.9 & 31.9\tiny$\pm$3.1 & \textbf{16.2\tiny$\pm$1.1} \\
                         & $\lambda{>}0$        & \textbf{13.9\tiny$\pm$2.1} (0.3) & \textbf{11.0} (0.5) & \textbf{10.2\tiny$\pm$1.8} (0.1) & \textbf{14.4} (0.3) & \textbf{11.8\tiny$\pm$0.4} (0.1) & 16.4\tiny$\pm$0.9 (0.5) \\
\midrule
\multirow{2}{*}{FNO}     & $\lambda{=}0$        & 24.5\tiny$\pm$2.0 & \textbf{17.2\tiny$\pm$3.8} & 14.9\tiny$\pm$3.5 & \textbf{14.0\tiny$\pm$2.4} & 24.4\tiny$\pm$4.2 & \textbf{14.4\tiny$\pm$0.9} \\
                         & $\lambda{>}0$        & \textbf{10.9\tiny$\pm$2.5} (0.1) & 18.3\tiny$\pm$2.8 (0.5) & \textbf{9.0\tiny$\pm$2.2} (0.1) & 15.4\tiny$\pm$2.3 (0.5) & \textbf{7.8\tiny$\pm$1.2} (0.1) & 16.0\tiny$\pm$1.0 (0.3) \\
\midrule
\multirow{2}{*}{ViT}     & $\lambda{=}0$        & 21.0\tiny$\pm$3.4 & 18.6\tiny$\pm$4.2 & 14.9\tiny$\pm$2.2 & \textbf{13.6\tiny$\pm$3.5} & 23.8\tiny$\pm$2.3 & \textbf{14.0\tiny$\pm$1.6} \\
                         & $\lambda{>}0$        & \textbf{18.8\tiny$\pm$2.3} (0.3) & \textbf{17.4\tiny$\pm$0.5} (0.7) & \textbf{14.0\tiny$\pm$2.2} (0.1) & 17.0\tiny$\pm$4.8 (0.5) & \textbf{14.7\tiny$\pm$0.6} (0.3) & 15.9\tiny$\pm$1.2 (0.3) \\
\midrule
\multirow{2}{*}{Diffusion} & $\lambda{=}0$        & 154.4\tiny$\pm$16.4 & 91.2\tiny$\pm$4.2 & 145.2\tiny$\pm$25.4 & 99.0\tiny$\pm$5.0 & \textbf{50.7\tiny$\pm$4.1} & \textbf{51.1\tiny$\pm$3.4} \\
                          & $\lambda{>}0$        & \textbf{108.5\tiny$\pm$19.0} (0.1) & \textbf{78.1\tiny$\pm$4.0} (0.1) & \textbf{89.6\tiny$\pm$9.1} (0.7) & \textbf{70.5\tiny$\pm$7.4} (0.1) & 80.6\tiny$\pm$6.4 (0.1) & 61.5\tiny$\pm$8.9 (0.3) \\
\bottomrule
\end{tabular}
\end{table}

On val, CoFINN reduces the error on drag coefficient in $11/12$ (model, range) cells; the only regression is diffusion on Broad. CNNFoil drops $57.3\%\to13.9\%$ Narrow ($4.1\times$), $40.6\%\to10.2\%$ Medium ($4.0\times$), $31.9\%\to11.8\%$ Broad ($2.7\times$). FNO drops $24.4\%\to7.8\%$ Broad ($3.1\times$). ViT drops $21.0\%\to18.8\%$ Narrow, $14.9\%\to14.0\%$ Medium, $23.8\%\to14.7\%$ Broad. Diffusion drops $154.4\%\to108.5\%$ Narrow, $145.2\%\to89.6\%$ Medium ($C_D$) and $91.2\%\to78.1\%$ Narrow, $99.0\%\to70.5\%$ Medium ($C_L$); regresses to $80.6\%$ Broad ($C_D$) and $61.5\%$ Broad ($C_L$). Lift on val: CNNFoil drops $22.7\%\to11.0\%$ Narrow, $18.8\%\to14.4\%$ Medium, $16.2\%\to16.4\%$ Broad; FNO and ViT regress within $\pm 2\%$ across all three ranges.

\SubHead{Flow Field Visualizations}

\begin{figure}[ht]
\centering
\includegraphics[width=\textwidth]{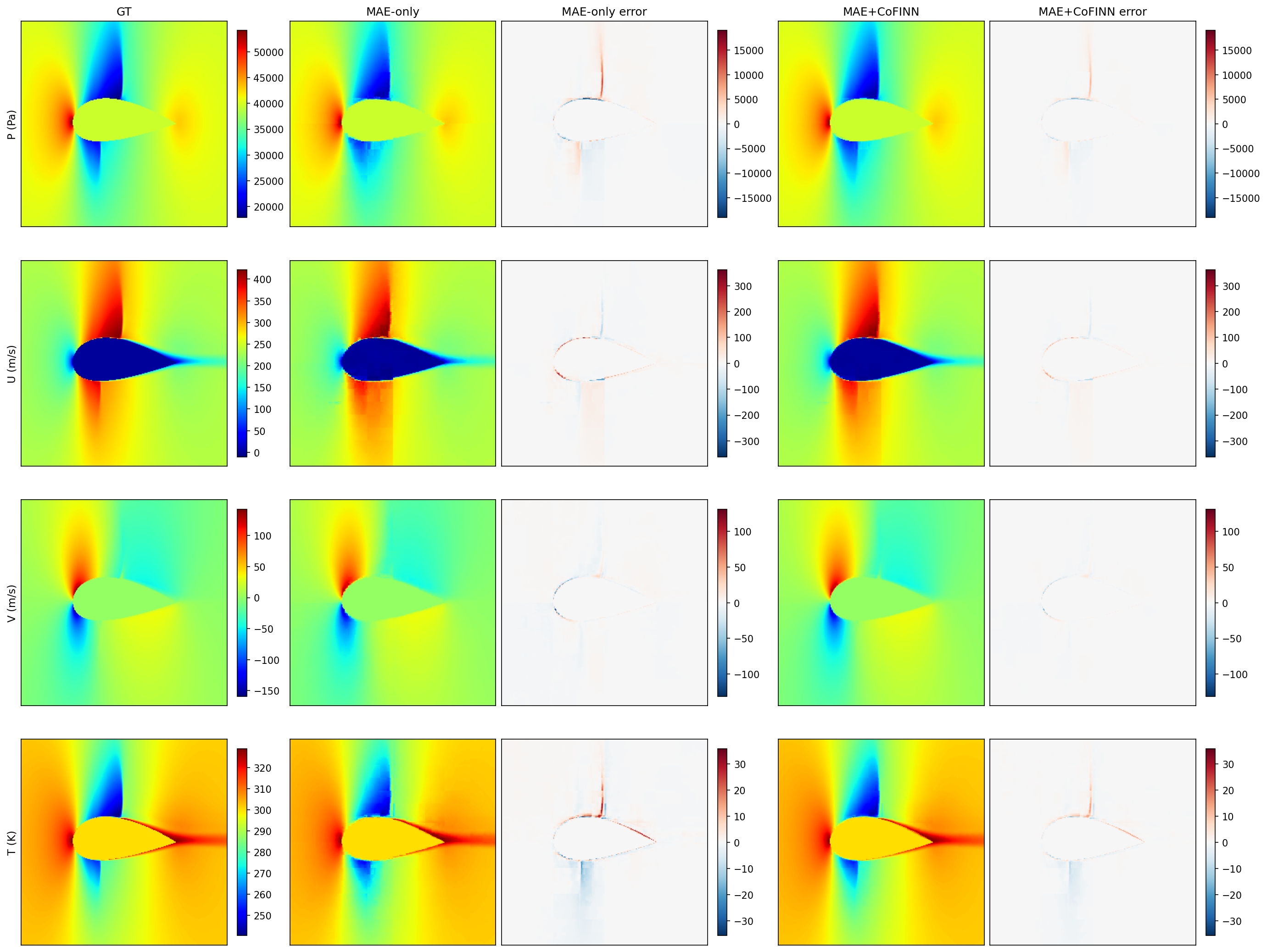}
\caption{\label{fig:flow-aoa0}Flow-field comparison for NACA~2421 at $\alpha=0^\circ$ (CNNFoil, Single training range). Columns: ground truth, MAE-only prediction, MAE-only error (signed), MAE+CoFINN prediction, MAE+CoFINN error (signed). Rows: pressure, $u$-velocity, $v$-velocity, temperature. Error colorbars are symmetric and shared between the two methods to enable direct visual comparison.}
\end{figure}

\begin{figure}[ht]
\centering
\includegraphics[width=\textwidth]{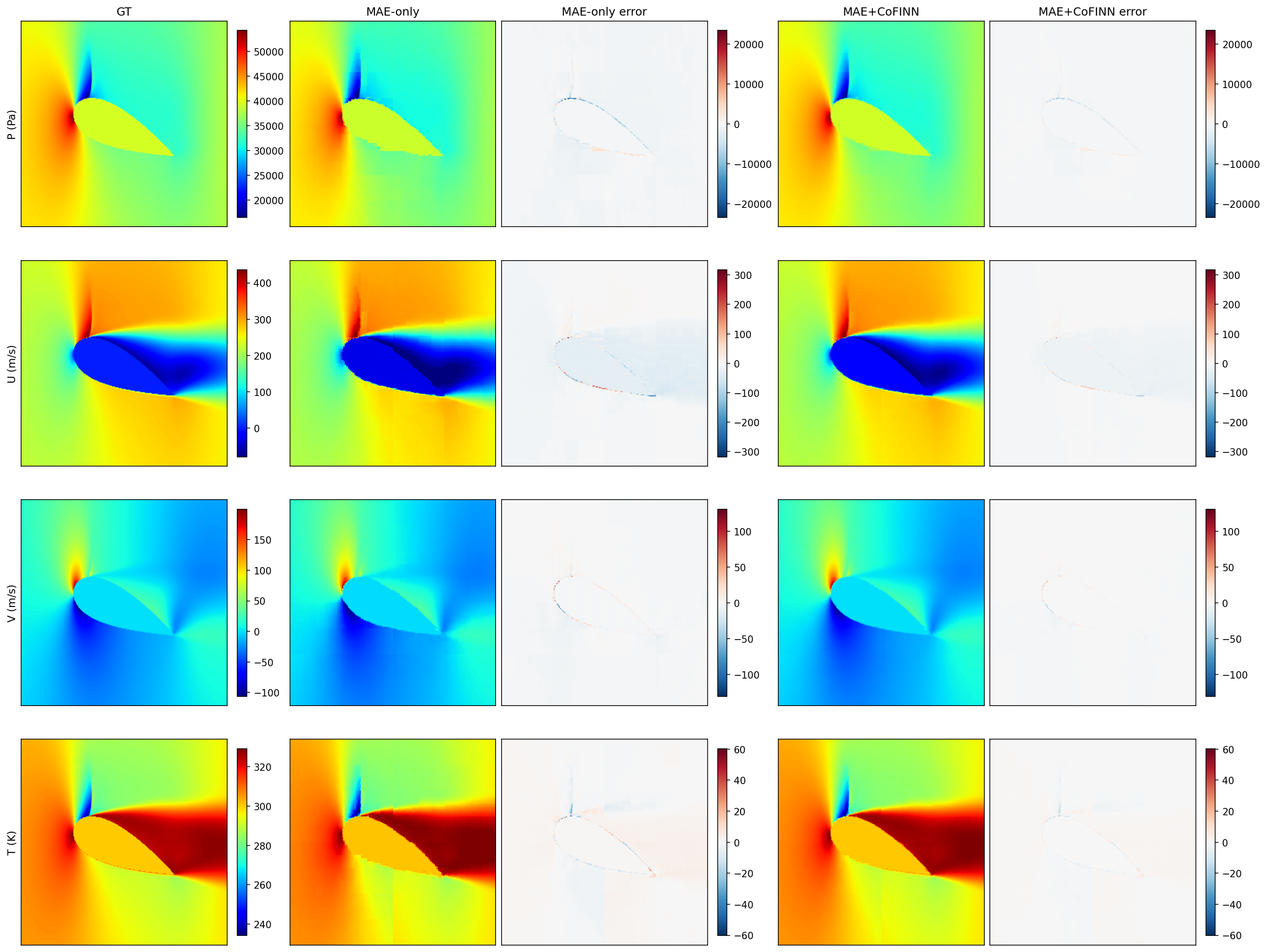}
\caption{\label{fig:flow-aoa12}Flow-field comparison for NACA~2421 at $\alpha=12^\circ$ (CNNFoil, Full training range). Same layout as Fig.~\ref{fig:flow-aoa0}.}
\end{figure}

Figures~\ref{fig:flow-aoa0} and~\ref{fig:flow-aoa12} compare predicted flow fields against CFD ground truth for the NACA~2421 test airfoil at $\alpha=0^\circ$ and $\alpha=12^\circ$, showing all four primitive variables ($p$, $u$, $v$, $T$). Each figure is a 4-row $\times$ 5-column grid: rows are pressure ($p$), $u$-velocity, $v$-velocity, temperature ($T$); columns are ground truth, MAE-only prediction, MAE-only error, MAE+CoFINN prediction, MAE+CoFINN error. All predictions shown in these figures are prepared by the CNNFoil architecture. The MAE-only column reports the lowest-field-MAE checkpoint, and the MAE+CoFINN column the prediction obtained when conservation-flux supervision is added to the same architecture at the same training regime.

At $\alpha=0^\circ$, both models successfully predicts general macroscopic flow features, including the attached velocity field and the pressure and temperature distributions associated with the shock waves. However, several key distinctions emerge: 

First, the CoFINN method significantly improves the prediction of shock wave location and strength. For this specific airfoil, shocks form on both the suction (upper) and pressure (lower) surfaces. While the baseline MAE model adequately captures the suction-side shock—likely due to a higher prevalence of such features in the training dataset—it struggles significantly on the pressure side. By incorporating physical conservation laws, the CoFINN framework effectively overcomes this data bias, accurately resolving the pressure-side shock.

Second, the error distributions confirm an improvement in resolving shock sharpness. In purely data-driven approaches, shock locations act as a primary source of high-magnitude local error. The enhanced shock-capturing capability in the CoFINN predictions can be attributed to the integration of the HLLC flux calculation within the physics-informed loss function, which naturally accommodates flow discontinuities.

Third, the MAE-only error contours (particularly for pressure) exhibit a non-physical, wavy artifact downstream of the pressure-side shock, indicative of numerical smearing as the data-driven model attempts to reconstruct the field. In contrast, CoFINN effectively suppresses these non-physical local oscillations.

Finally, pure CNN-based architectures traditionally struggle with thin, viscous boundary layers due to steep gradients and limited spatial resolution near the wall. While CoFINN yields marginal visual improvements in this region, the near-wall data remains fundamentally insufficient for reliable surface integration.  Recognizing this inherent limitation, our methodology deliberately bypasses the near-wall inaccuracies by employing a large, rectangular far-field control volume.

%The MAE-only error panels expose residual pressure ripples downstream of the trailing edge; MAE+CoFINN reduces these residuals across all four channels, with the largest relative improvement on the velocity components where momentum conservation is most directly enforced.

At $\alpha=12^\circ$, a strong shock forms on the suction side, followed by a region of massive flow separation. While both the MAE-only and CoFINN predictions successfully capture the macroscopic shock location and post-shock pressure recovery, the MAE-only error panels reveal that dominant residuals are heavily concentrated around the shock front and throughout the separated wake region.

CoFINN substantially mitigates these residuals, particularly within the wake. In this region, satisfying the $x$-momentum flux balance is critical for the physical fidelity of both the streamwise velocity ($u$) and pressure fields. This local improvement translates directly into highly accurate drag predictions, as demonstrated in the per-angle drag breakdown (Table~\ref{tab:per-aoa-cd}). Specifically, CoFINN reduces the baseline $C_D$ error ($\epsilon_{C_D}$) at $\alpha=12^\circ$ from $24.5\%$ down to $3.8\%$ ($\lambda=0.3$), confirming that enforcing the control-volume momentum integral effectively concentrates field improvements precisely where complex flow behaviors dictate aerodynamic forces.

A spatially resolved comparison reveals why: the CoFINN loss acts non-uniformly across the flow domain. In the wake region behind the airfoil, where momentum flux gradients are largest, CoFINN-trained models reduce pressure error by $30$--$45\%$ relative to the baseline, with over $75\%$ of wake pixels showing improvement. In contrast, the near-body and far-field regions show $15$--$45\%$ \emph{increased} error. The improvement concentrates on the momentum-carrying variables ($p$, $u$) rather than temperature or crossflow velocity ($v$), consistent with the HLLC flux calculation~(Eq.~\eqref{eq:riemann-flux}) in which pressure and normal velocity dominate the momentum terms while temperature enters only indirectly through density and sound speed estimation.

This spatial selectivity explains the MAE--$C_D$ tradeoff observed throughout: the CoFINN loss redirects model capacity from uniformly minimizing pixel-wise error toward satisfying flux conservation in the wake---precisely the region whose momentum integral determines drag (Eq.~(S17) ). %~\eqref{supp-eq:lift-drag-integral}). 
The far-field pixels, which dominate the pixel count but contribute little to force integration, lose accuracy in this re-balancing.

Beyond the shock structures and separated wakes, a critical aerodynamic distinction between the models emerges at the leading-edge stagnation point. In both the $\alpha=0^\circ$ and $\alpha=12^\circ$ configurations, the MAE-only error contours for pressure and temperature exhibit concentrated, highly localized artifacts precisely where the flow stagnates against the airfoil.  In contrast, the CoFINN framework noticeably suppresses these stagnation artifacts. Even as the stagnation point migrates to the lower surface at $\alpha=12^\circ$, CoFINN maintains higher fidelity at this critical flow impact zone. From an aerodynamic perspective, the stagnation point anchors the total pressure and total enthalpy for the entire flow field. By enforcing conservation laws across the domain, the CoFINN loss function inherently respects these maximum state variables.

\SubHead{Assessment of aerodynamic performance predictions}

\begin{figure}[ht]
\centering
\includegraphics[width=0.7\textwidth]{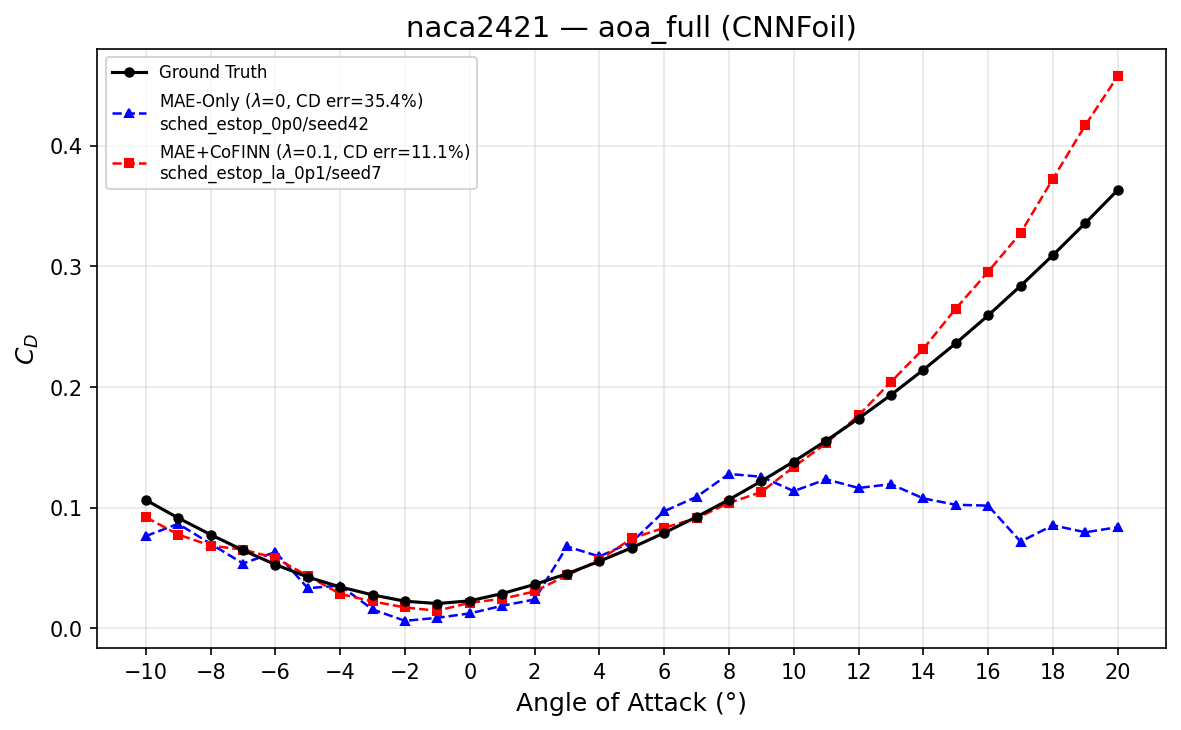}
\caption{\label{fig:cd-aoa-naca2421}$C_D$ vs.\ angle of attack for NACA~2421 (eval partition, Full range, CNNFoil). Black: CFD ground truth, blue: MAE-only baseline ($\lambda{=}0$), red: CoFINN (best $\lambda{>}0$). The baseline misses negative-angle behavior while CoFINN tracks the CFD curve more closely.}
\end{figure}

Figure~\ref{fig:cd-aoa-naca2421} illustrates the predicted drag coefficient ($C_D$) across the full range of angles of attack for the NACA 2421 airfoil. The ground truth data demonstrates the expected parabolic drag polar, with a sharp increase in drag at higher angles ($\alpha > 10^\circ$) due to the onset of flow separation and massive wake formation.The baseline MAE-only model ($\lambda=0$) struggles significantly across the domain. Most notably, it completely fails to capture the physical rise in drag at high angles of attack, severely underpredicting $C_D$ in the separated flow regimes.

In contrast, the CoFINN model ($\lambda=0.1$) yields a substantial improvement. The physics-informed loss function enables the network to accurately track the steep increase in drag well into the high angle-of-attack regime. While there is a slight overprediction at the most extreme angles ($\alpha > 14^\circ$), CoFINN successfully captures the macroscopic drag trends. This aligns with our earlier spatially resolved observations: by enforcing $x$-momentum conservation, CoFINN correctly resolves the momentum deficit in the wake, directly translating to vastly improved drag integration at the far-field control volume.

\begin{figure}[ht]
\centering
\includegraphics[width=0.7\textwidth]{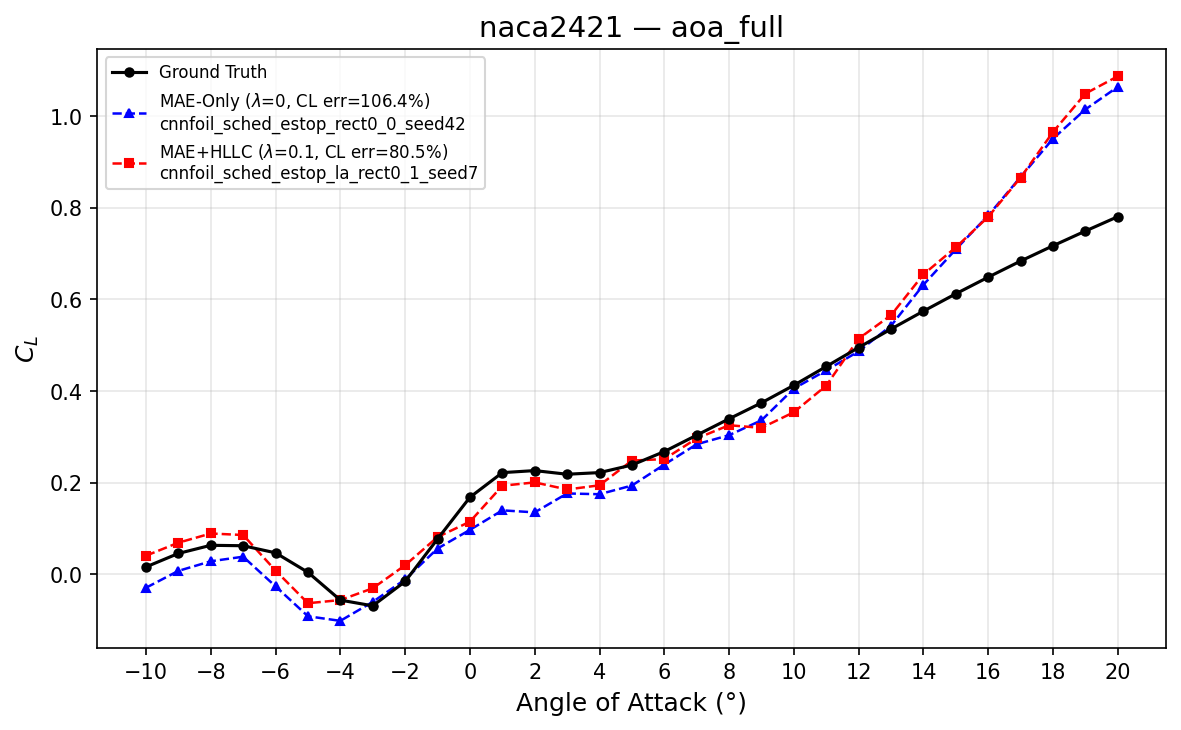}
\caption{\label{fig:cl-aoa-naca2421}$C_L$ vs.\ angle of attack for NACA~2421 (eval partition, Full range, CNNFoil). Black: CFD ground truth, blue: MAE-only baseline ($\lambda{=}0$), red: CoFINN (best $\lambda{>}0$). The baseline misses negative-angle behavior while CoFINN tracks the CFD curve more closely.}
\end{figure}

Following the pronounced improvements observed in the drag polar, Figure~\ref{fig:cl-aoa-naca2421} shows predicted $C_L$ across all 31 angles for the NACA~2421 test airfoil on the Full training range, comparing the best MAE-only baseline ($\lambda{=}0$) and best CoFINN ($\lambda{>}0$) experiment. At the attached and mildly separated flow regimes between $-10^\circ \leq \alpha \leq 10^\circ $,  the physics-informed loss successfully improves the drag predictions compared to CNNFoil. 

However, a contrast to the drag predictions emerges in the deep post-stall regime ($\alpha > 12^\circ$). While CoFINN successfully captured the sharp drag rise associated with massive separation, both the baseline and CoFINN models fail to predict the corresponding stall drop-off in lift. Instead, the networks extrapolate a nearly linear, non-physical increase in $C_L$.

This discrepancy can be attributed to a gradient scaling pathology within the unweighted, multi-objective loss formulation. Aerodynamically, the streamwise velocity ($u$) and its associated gradients are intrinsically larger in magnitude than the crossflow velocity ($v$). Because the current loss function aggregates all physical residuals into a single, equally-weighted metric, the massive $x$-momentum errors generated in the wake completely dominate the total loss metric.

As a result, the gradient descent optimizer preferentially dedicates the network's learning capacity to minimizing these dominant $x$-momentum (drag-governing) errors. The numerically smaller $y$-momentum (lift-governing) residuals are overshadowed and optimized far less aggressively. Consequently, while CoFINN successfully enforces local momentum conservation in the drag direction, it lacks the gradient leverage required to overcome the base network's linear bias and force the transition to a globally stalled flow topology.

We additionally evaluate a conditional diffusion model based on the Bastek et al.~\cite{bastek_physics-informed_2024} U-Net architecture. The implementation details and brief discussion of results are given in the %section~ \ref{supp-sec:diffusion-model} in the 
supplementary text.

\Head{Discussion}
\label{sec:conclusion}

We introduced CoFINN, a conservation-flux-informed training framework that embeds finite-volume conservation principles directly into neural-network training by interpreting CNN outputs as structured control-volume grids and penalizing flux imbalance through an HLLC Riemann solver. Unlike conventional data-driven surrogate models that optimize only field similarity, CoFINN explicitly guides the network toward physically admissible solutions consistent with the governing conservation laws of compressible flow.

Across transonic airfoil test cases ($M_\infty=0.7$, $Re=6 \times 10^6$), the proposed framework consistently improved aerodynamic force prediction accuracy, particularly under physically challenging flow conditions involving strong shocks and high-angle-of-attack effects. Drag prediction error was reduced by up to 34\% at extreme angles of attack and by approximately 15\% on average across the full test set. Lift predictions exhibited even more consistent improvement across all training regimes. These results suggest that enforcing conservation behavior during training improves not only local field consistency but also the accuracy of integrated physical quantities derived from the predicted flow fields.

An important observation is that the benefits of CoFINN become more pronounced as the amount of training data decreases. In the single-angle-of-attack training regime, the CoFINN training improved drag prediction accuracy by more than 34\%, indicating that the conservation loss acts as an effective physical regularizer. Rather than merely memorizing spatial correlations present in the training data, the network is encouraged to learn flow structures that satisfy physically meaningful transport behavior. This observation supports the broader hypothesis that embedding numerical physics into learning architectures can improve generalization capability in data-limited regimes.

The proposed methodology differs fundamentally from traditional Physics-Informed Neural Networks (PINNs). Classical PINNs typically enforce differential-equation residuals at collocation points through automatic differentiation, which can become computationally expensive for high-resolution flow-field prediction problems. In contrast, CoFINN adopts a finite-volume perspective consistent with modern CFD methodology. By evaluating numerical fluxes between neighboring cells using a Riemann solver, conservation is enforced in a discretized integral sense rather than through pointwise PDE residual minimization. This provides a direct connection between neural-network training and established CFD discretization principles.

Another important contribution of the present framework is the use of far-field control-volume integration for aerodynamic force estimation. CNN-based flow predictors often struggle near solid boundaries because geometric representation and near-wall gradients are difficult to resolve accurately on image-like grids. By computing forces from conservation balances over larger control volumes, the proposed method reduces sensitivity to near-wall prediction inaccuracies while remaining physically consistent with momentum conservation principles.

The computational overhead introduced by the conservation loss remains modest. While HLLC flux evaluation increases training cost by approximately 20\%, inference cost is unchanged relative to baseline CNN models. Prediction times remain on the order of 50 ms per sample on an NVIDIA RTX 3090 GPU, corresponding to nearly three orders of magnitude acceleration compared to high-fidelity CFD simulations requiring several CPU-hours per case. This demonstrates that physically informed training can significantly improve physical fidelity without compromising the practical efficiency advantages of neural-network surrogates.

Despite these improvements, several important limitations remain. The $256 \times 256$ structured representation does not adequately resolve the thin viscous boundary layer near the airfoil surface. Consequently, the Euler-based HLLC flux primarily constrains inviscid momentum transport and has limited influence on near-wall viscous phenomena such as skin friction and boundary-layer separation. This explains why improvements are strongest at extreme angles of attack dominated by large-scale inviscid transport and shock physics, whereas gains are more modest in flow regimes where viscous effects are comparatively important.

The conservation loss also cannot function as a standalone training objective. Training with the CoFINN loss alone ($\lambda=1.0$) leads to unstable optimization and divergence, indicating that data supervision remains necessary to anchor the solution manifold. In the present formulation, the CoFINN loss therefore acts as a physically motivated regularizer rather than a replacement for supervised learning. Furthermore, the optimal weighting between data loss and conservation loss varies across flow conditions, suggesting that adaptive or dynamically weighted strategies may further improve robustness.
The current study also remains limited to two-dimensional compressible flow at a single Mach and Reynolds number combination. Although the framework itself is architecture-agnostic and compatible with CNNs, FNOs, Vision Transformers, and diffusion-based models, additional validation across broader aerodynamic regimes and three-dimensional geometries is required before conclusions can be generalized.
Nevertheless, the proposed framework establishes a broader conceptual direction for physics-informed machine learning. The key idea of interpreting neural-network outputs as finite-volume discretizations is not restricted to airfoil aerodynamics or to the HLLC solver specifically. Any conservation-law-governed physical system defined on structured grids may potentially benefit from similar conservation-flux supervision using appropriate numerical flux formulations.

Several extensions of the present work appear especially promising. First, the framework is not restricted to Godunov-type Riemann solvers; central schemes such as JST or alternative flux formulations may provide improved stability or computational efficiency. Second, higher-order spatial reconstruction methods and slope limiters could be incorporated into the conservation-loss calculation to improve consistency with higher-order CFD discretizations. Third, explicit inclusion of viscous flux terms would extend the method beyond primarily inviscid conservation enforcement.

The inclusion of viscous physics, however, presents additional challenges. Turbulent viscous fluxes require estimation of quantities such as turbulent viscosity and potentially the inclusion of auxiliary turbulence-model transport equations. Moreover, accurate representation of near-wall viscous physics requires substantially finer geometric resolution near solid boundaries than conventional CNN image grids can provide. Future extensions using graph neural networks or mesh-aware architectures may therefore provide a more suitable framework for accurately resolving complex near-wall flow behavior while retaining the conservation-flux-informed training philosophy introduced in this work.

\bibliographystyle{plainnat}
\bibliography{bibliography}

\end{document}